%% file: main.tex
\documentclass[pmlr]{jmlr}
\jmlrproceedings{PMLR}{Preprint}

\RequirePackage{graphicx}
\usepackage{longtable}

\usepackage{breqn}
\usepackage{tikz}
\usepackage{xcolor}
\usepackage{booktabs}
\usepackage{threeparttable}
\usepackage{float}
\usepackage{pdflscape}

\usepackage[textsize=tiny]{todonotes}

\usepackage{mathtools}

\usepackage{hyperref}
\usepackage[capitalize,noabbrev]{cleveref}

\input{Utilities/math_commands}
\input{Utilities/diagram_settings}

\title[Deep Kernel Learning for Glaucoma]{Deep Kernel Learning for Stratifying Glaucoma Trajectories}

\author{%
  \Name{Bruce Rushing}
  \Email{ejf9db@virginia.edu}\\
  \addr Research Computing, University of Virginia
  \AND
  \Name{Angela Danquah}
  \Email{aab5zd@virginia.edu}\\
  \addr Research Computing, University of Virginia
  \AND
  \Name{Alireza Namazi}
  \Email{mez4em@virginia.edu}\\
  \addr School of Data Science, University of Virginia
  \AND
  \Name{Arjun Dirghangi}
  \Email{ad3bb@uvahealth.org}\\
  \addr School of Medicine, University of Virginia
  \AND
  \Name{Heman Shakeri}
  \Email{hs9hd@virginia.edu}\\
  \addr School of Data Science, University of Virginia
}

\begin{document}

\maketitle

\begin{abstract}
Effectively stratifying patient risk in chronic diseases like glaucoma is a major clinical challenge. Clinicians need tools to identify patients at high risk of progression from sparse and irregularly-sampled electronic health records (EHRs).
We propose a novel deep kernel learning (DKL) architecture that leverages a Gaussian Process (GP) backend. The GP's kernel is defined by a transformer-based feature extractor applied to clinical-BERT embeddings to model glaucoma patient trajectories from multimodal EHR data.
Our method successfully identifies three clinically distinct patient subgroups. Crucially, the model learns to decouple disease progression from current severity, identifying a high-risk group with a worsening trajectory despite having better average visual acuity than a second, stably poor group. This reveals that the model learns to identify progression risk rather than just the current disease state. 
 This ability to stratify patients based on their risk trajectory progression offers a powerful tool for clinical decision support, enabling targeted interventions for high-risk individuals and improving the management of glaucoma care.
\end{abstract}

\input{Sections/introduction}
\input{Sections/related}
\input{Sections/preliminaries}
\input{Sections/disease_trajectories}
\input{Sections/experiments}
\input{Sections/discussion}


\bibliography{mlhc}

\newpage
\appendix

\input{Sections/appendix_a}
\input{Sections/appendix_b}
\input{Sections/appendix_c}

\input{Sections/appendix_d}

\end{document}

%% file: Utilities/math_commands.tex
\newcommand{\mathbi}[1]{\textbf{\em #1}}

\newcommand{\randvecfunction}{\mathbi{f}}
\newcommand{\randX}{\mathit{X}}
\newcommand{\randx}{\mathit{x}}
\newcommand{\randvecX}{\mathbi{X}}
\newcommand{\randvecx}{\mathbi{x}}

\newcommand{\randy}{\mathit{y}}

\newcommand{\randvecy}{\mathbi{y}}
\newcommand{\randvecw}{\mathbi{w}}

\newcommand{\expect}{\mathbb{E}}
\newcommand{\priorcovariance}{\Sigma_{p}}
\newcommand{\normal}[2]{\mathcal{N}\left(#1, #2\right)}
\newcommand{\condnormal}[3]{\mathcal{N}\left(#1 \penalty10000\mid #2,\penalty0\allowbreak #3\right)}
\newcommand{\gp}[2]{\mathcal{GP}(#1, #2)}

\newcommand{\features}{\mathcal{X}}
\newcommand{\targets}{\mathcal{Y}}
\newcommand{\data}{\mathcal{D}}
\newcommand{\dataset}{\{(\randvecx_{i}, \randy_{i})\}_{i=1,\dots,n}}

\newcommand{\matrixA}{\mathbf{A}}

\newcommand{\matrixX}{\mathbf{X}}
\newcommand{\matrixZ}{\mathbf{Z}}
\newcommand{\identity}{\mathbf{I}}

\newcommand{\zerovec}{\mathbf{0}}

\newcommand{\transpose}[1]{#1^{\intercal}}
\newcommand{\reals}[1]{\mathbb{R}^{#1}}
\newcommand{\normltwo}[2]{\Vert #1 - #2 \Vert^{2}}

\newcommand{\bigo}[1]{\mathcal{O}(#1)}

%% file: Utilities/diagram_settings.tex
\usetikzlibrary{shapes.geometric, arrows.meta, positioning}

\definecolor{darkblue}{RGB}{25,25,112}
\definecolor{lightblue}{RGB}{173,216,230}
\definecolor{mediumblue}{RGB}{135,206,250}
\definecolor{orange}{RGB}{255,140,0}

\tikzset{
    process/.style={
        rectangle, 
        draw=darkblue, 
        fill=mediumblue!60,
        align=center,
        inner sep=4pt,
        rounded corners=3pt,
        line width=0.7pt,
        font=\footnotesize
    },
    data/.style={
        trapezium,
        trapezium left angle=75,
        trapezium right angle=105,
        draw=darkblue,
        fill=lightblue!70,
        align=center,
        inner sep=4pt,
        line width=0.7pt,
        font=\footnotesize
    },
    decision/.style={
        diamond,
        draw=darkblue,
        fill=mediumblue!60,
        align=center,
        inner sep=3pt,
        aspect=2,
        line width=0.7pt,
        font=\footnotesize
    },
    output/.style={
        rectangle,
        draw=orange!80!black,
        fill=orange!30,
        align=center,
        inner sep=4pt,
        rounded corners=3pt,
        line width=0.8pt,
        font=\footnotesize
    },
    arrow/.style={
        ->,
        >=Stealth,
        line width=0.8pt,
        darkblue
    },
    sidenote/.style={
        font=\scriptsize\itshape,
        text=gray!60!black,
        align=left
    }
}

%% file: Sections/introduction.tex
\section{Introduction}

Irregular time-series data is a major problem when analyzing longitudinal data sources such as electronic health records (EHRs) \citep{che2018recurrent, shukla2019modeling, zhang2022improving}. Entries with features that vary with visit and that occur at different intervals make it extremely challenging to do forecasting from such data \citep{karami2024tee4ehr, si2021deep}. This presents even more problems when forecasting models are leveraged to identify different patient clusters; clinicians are often interested in diagnosing when patients might be primed for improvement or downward health trends based on their history of observations, so utilizing models to forecast relevant diagnostic criteria and then cluster patients based on those models' representations can be extraordinarily useful for clinical practice \citep{schulam2015clustering, feuerriegel2021analyzing}. The upshot for medical practitioners on having forecasting methods that can be used to identify patient clinical trajectories prior to negative trajectories is improved predictive methods that would enable better therapeutic treatments \citep{placido2023deep, liu2019representation}.

In addition to the sparsity of features and observations over time, there are two further problems that confound time-series forecasts for EHRs. First, the data is high-dimensional---especially when including natural language features like clinical notes \citep{rajkomar2018scalable, alsentzer2019publicly}. This can be problematic for traditional methods in irregular time-series forecasts and patient clustering strategies like Gaussian Processes (GP) that suffer high time and memory complexity in dataset size and dimensionality \citep{williams2006gaussian, schulam2015clustering}. Second, if the goal of the forecast is to identify patient trajectories, then the number of samples of records for most patients will be low. State-of-the-art (SOTA) methods for EHR forecasting that leverage large language models are anything but sample efficient, making them poor candidates for effective trajectory modeling based on limited patient records \citep{yang2022large, rasmy2021med}.

We offer a hybrid solution that employs pre-trained language model embeddings along with a recurrent or transformer-based deep kernel learning (DKL) algorithm to simultaneously address the problems of data irregularity, sparsity, high-dimensionality, and patient-relative low sample size \citep{wilson2016deep, alsentzer2019publicly}. Notably, our method does not rely upon imputation with temporal process points that are common for addressing data irregularity. Using the SOURCE glaucoma dataset of EHRs, we successfully forecast irregular time-series data for predicting glaucoma patient visual acuity loss. Our method either matches or outperforms most traditional recurrent neural network and transformer-based forecasting methods. We use the GP backend of our DKL algorithm to identify three clinically distinct patient trajectories that reveal a critical insight: patients with moderate vision but high trajectory variance represent high-risk group requiring intensive monitoring, despite those patients' better vision compared to chronically poor but stable patients. This offers expanded opportunities to identify high-risk patients by decoupling current disease severity from future progression risk, enabling targeted interventions for those most likely to experience vision loss.

Our main contributions are as follows:

\begin{itemize}
    \item A novel hybrid architecture that pre-processes EHR natural language features through clinical-BERT embeddings and then trains a DKL algorithm with a transformer-encoder feature extractor to predict patient visual acuity loss.
    \item We demonstrate with experiments better hold-out test metrics on the SOURCE dataset using our method compared to standard time-series recurrent neural networks and transformers.
    \item We show that our method identifies three clinically meaningful patient subgroups by decoupling current disease state from progression risk, with the models providing calibrated uncertainty estimates---achieving 53.06\% accuracy within 0.1 logMAR.
    \item Our framework provides calibrated uncertainty quantification that enables risk-aware clinical decision support, with high-variance predictions correctly identifying patients with volatile disease courses rather than just reflecting model uncertainty.
\end{itemize}

\noindent \textbf{Generalizable Insights}: We provide probabilistic forecasts that can aid in handling uncertainty in healthcare. By pairing a transformer encoder with a Gaussian Process head, probabilistic deep learning can ingest irregularly sampled, multimodal records, such as  unstructured clinical notes, and return calibrated uncertainty that distinguishes irreducible patient variability from model ignorance. Our modeling approach shows that a patient's current disease severity poorly predicts future trajectory: unsupervised clustering of the learned representations recovers progression subtypes consistent with landmark glaucoma trials, without access to gold-standard structural or functional measurements. We can leverage the uncertainty provided by our method to flag volatile patients for further monitoring. By quantifying uncertainty honestly over messy longitudinal data and decoupling disease state from progression risk, this method can apply to chronic conditions from diabetes to heart failure, where irregular follow-up and heterogeneous trajectories are the rule.

%% file: Sections/related.tex
\section{Related Work} \label{sec:related}

Managing irregular time-series data has been an ongoing project in machine learning, with recent work focusing on leveraging deep learning architectures. While feedforward, convolutional, and recurrent networks commanded the most attention \citep{lim2021time, ling2023comprehensive}, transformers \citep{vaswani2017attention} now predominate. Custom transformer architectures for time-series include modified positional encoders and attention mechanisms \citep{wen2023transformers}. One prominent line of research involves adjusting the positional encoder with a Hawkes process to integrate interval information \citep{zhang2020self, zuo2020transformer}, while sophisticated embeddings can further improve performance \citep{mei2022transformer}. Recent work converts irregular time-series into line-graphs for vision transformer processing \citep{li2023time, dosovitskiy2021an}.

Machine learning for EHRs has advanced significantly \citep{si2021deep, shickel2017deep}. Early Probabilistic Subtyping Models addressed EHR irregularity and sparsity \citep{schulam2015clustering}, and the subsequent application of deep learning has led to more powerful models. GRU-D \citep{che2018recurrent} handles irregular data by explicitly modeling informative missingness---missing data salient for predictions. Recurrent networks address high-dimensional EHR data \citep{rajkomar2018scalable}, despite reproducibility challenges \citep{si2021deep}. Attention has increasingly focused on transformers: transformer encoders with point processes can successfully classify irregular EHR data \citep{karami2024tee4ehr}, and multi-modal EHR data has been integrated to produce better predictions \citep{zhang2022improving}. For glaucoma, EHR-based surgery prediction uses regression, decision trees, and neural networks \citep{wang2023prediction}, including word embedding approaches \citep{wang2022deep}. 

Prior EHR work emphasizes regression or classification. However, clinicians need to group patient clinical history into trajectories for predicting disease progression \citep{yohannan2020assessing, feuerriegel2021analyzing}. EHR-based trajectory modeling remains challenging. Gaussian processes address this challenge \citep{williams2006gaussian, hensman2013gaussian}. Disease Trajectory Maps \citep{schulam2016disease} use Gaussian processes with low-dimensional clinical markers to model autoimmune progression through latent spaces. More recent work has utilized transformers combined with probabilistic models to measure disease progression and cluster trajectories \citep{qiu2025deep}. Their model uses a transformer-based variational autoencoder with Gaussian mixture priors to jointly model patient clustering and survival times from diagnosis sequences. While this approach successfully identifies disease subgroups with divergent trajectories, it focuses primarily on structured diagnosis codes rather than leveraging the rich semantic information in clinical notes and uses mixture models instead of Gaussian processes for uncertainty quantification.

%% file: Sections/preliminaries.tex
\section{Preliminaries}

We denote sets as $\mathcal{A}$, random variables and assignments as $\randX$ and $\randx$, random vectors and assignments as $\randvecX$ and $\randvecx$, and matrices as $\matrixA$ and $\Sigma$. Given dataset $\data = \dataset$ with features $\features$ and targets $\targets$, Bayesian inference computes the posterior predictive distribution for parametric model $f: \features \to \targets$ characterized by parameters $\randvecw$:

\begin{equation} \label{eq:posteriorpred}
    p(\randy \mid \randvecx, \data) = \int p(\randy \mid \randvecx; \randvecw) p(\randvecw \mid \data) d\randvecw
\end{equation}

\noindent where $p(\randvecw \mid \data) \propto p(\data \mid \randvecw)p(\randvecw)$. For Bayesian linear regression, we assume a likelihood $\condnormal{\randy}{\transpose{\randvecx}\randvecw}{\sigma^{2}}$ and a zero-mean Gaussian prior on the weights, $\randvecw \sim \normal{\zerovec}{\priorcovariance}$. Let $\matrixX$ and $\randvecy$ denote the stacked features and targets from $\data$. For nonlinear mappings, we project inputs $\phi: \features \to \reals{d}$ yielding $f(\randvecx; \randvecw) \mapsto \transpose{\phi(\randvecx)}\randvecw$. With design matrix $\Phi$, the posterior becomes:

\begin{dmath}
    p(\randy \mid \randvecx, \data) = \condnormal{\randy}{\transpose{\phi_{\randvecx}}\priorcovariance\Phi (K + \sigma^{2}\identity)^{-1}\randvecy}{\transpose{\phi_{\randvecx}}\priorcovariance\phi_{\randvecx} - \transpose{\phi_{\randvecx}}\priorcovariance\Phi(K + \sigma^{2}\identity)^{-1}\transpose{\Phi}\priorcovariance\phi_{\randvecx}}
\end{dmath}

\noindent where $\phi_{\randvecx} = \phi(\randvecx)$ and $K = \transpose{\Phi}\priorcovariance\Phi$. 

Since $\phi$ and $\Phi$ only occur in inner products with $\priorcovariance$, we define kernel $k(\randvecx, \randvecx^{\prime}) = \priorcovariance^{1/2} \phi(\randvecx) \cdot \priorcovariance^{1/2} \phi(\randvecx^{\prime})$, yielding a Gaussian process (GP) with mean $m(\randvecx) \mapsto \expect[f(\randvecx)]$:

\begin{equation}
    f(\randvecx) \sim \gp{m(\randvecx)}{k(\randvecx, \randvecx^{\prime})}
\end{equation}

\noindent The common squared exponential kernel is:

\begin{equation}
    k_{se}(\randvecx, \randvecx^{\prime}) = \exp\left(-\frac{\normltwo{\randvecx}{\randvecx^{\prime}}}{2l}\right)
\end{equation}

\noindent where $l$ is the length parameter. Applying the kernel element-wise over features $\matrixX$ and input points $\matrixX_{*}$, we get matrix $K(\cdot, \cdot)$. For noisy observations with zero mean prior, the prior is:

\begin{equation}
    \begin{bmatrix}
        \randvecy \\
        \randvecfunction_{*}
    \end{bmatrix} \sim \normal{\zerovec}{\begin{bmatrix}
        K(\matrixX, \matrixX) + \sigma^{2}\identity & K(\matrixX, \matrixX_{*}) \\
        K(\matrixX_{*}, \matrixX) & K(\matrixX_{*}, \matrixX_{*})
    \end{bmatrix}}
\end{equation}

\noindent and the posterior predictive is:

\begin{dmath}
    \randvecfunction_{*} \mid \matrixX_{*}, \data \penalty10000\sim \condnormal{\randvecfunction}{K(\matrixX_{*}, \matrixX)\Lambda^{-1}\randvecy}{K(\matrixX_{*}, \matrixX_{*}) - K(\matrixX_{*}, \matrixX)\Lambda^{-1}K(\matrixX, \matrixX_{*})}
\end{dmath}

\noindent where $\Lambda = K(\matrixX, \matrixX) + \sigma^{2}\identity$. Hyperparameter $l$ for our kernel $k_{se}$ is found via empirical Bayes, i.e. maximizing the marginal likelihood $\log p(\randvecy \mid \matrixX, l) = \log \condnormal{\randvecy}{\zerovec}{\Lambda} = -\frac{1}{2}\transpose{\randvecy}\Lambda^{-1}\randvecy - \frac{1}{2}\log \vert \Lambda\vert - \frac{n}{2}\log 2\pi$.

Computing the posterior requires inverting $\Lambda$ with $\bigo{n^{3}}$ time and $\bigo{n^{2}}$ space complexity. This makes utilizing GPs impractical for large datasets with high-dimensional features. Common solutions for this problem include: variational inference with inducing points for large datasets, and neural network dimensionality reduction for high-dimensional features.

For scalability, we partition features into training $\matrixX$, inducing point $\matrixZ$, and test $\matrixX_{*}$, with joint variational distribution $q(\randvecfunction_{\matrixX}, \randvecfunction_{\matrixZ}, \randvecfunction_{\matrixX_*}) \approx p(\randvecfunction_{\matrixX}, \randvecfunction_{\matrixX_{*}} \mid \randvecfunction_{\matrixZ})q(\randvecfunction_{\matrixZ})$. Our optimization target is to minimize the KL divergence between the variational and true posterior by the Evidence Lower Bound (ELBO) \citep{hensman2015scalable}:

\begin{equation} \label{eq:elbo}
    \mathcal{L}(q) = \expect_{q(\randvecfunction_{\matrixX})}[\log p(\randvecy \mid \randvecfunction_{\matrixX})] - D_{KL}(q(\randvecfunction_{\matrixZ}) \Vert p(\randvecfunction_{\matrixZ})) 
\end{equation}

\noindent To compute the first part of the ELBO, we utilize stochastic gradient descent over mini-batches.

For high-dimensional feature spaces, we map $h: \reals{d} \to \reals{k}$ ($k \ll d$) before computing kernels:

\begin{equation}
    k_{dkl\_se}(\randvecx, \randvecx^{\prime}) = \exp\left(-\frac{\normltwo{h(\randvecx)}{h(\randvecx^{\prime})}}{2l}\right)
\end{equation}

\noindent A neural network $h_{\theta}$ learns this mapping, yielding deep kernel learning \citep{wilson2016deep}. Parameters $\theta$ are treated as hyperparameters for the GP and jointly optimized with other hyperparameters via \ref{eq:elbo}. The increased hyperparameters risk overfitting \citep{ober2021promises}, which can be mitigated by mini-batching and additional regularization.

%% file: Sections/disease_trajectories.tex
\section{Disease Trajectories with Deep Kernel Learning}


EHRs induce challenges for time-series forecasting due to irregular sampling and missing features. GPs address irregular sampling by interpolating through prior and posterior estimates, and they handle missing values naturally by increased posterior variance, providing broader uncertainty estimates. The functions modeled by the multivariate normal distribution in a GP also enable modeling of different \textit{types} of the underlying generating population. This has made GPs especially useful in the medical domain for tracking population-level disease trajectories \citep{schulam2016disease} as the fitted posteriors allow for the clustering of entire trajectory functions.

High-dimensional EHR features pose computational challenges for GPs. First, they cannot effectively learn representations from clinical notes. Pre-trained embeddings, such as word2vec \citep{mikolov2013efficient}, provide usable features, but their high dimensionality create $\bigo{n^{3}}$ computational bottlenecks in kernel inversion. Second, time-series architectures like LSTM \citep{hochreiter1997long}, GRU \citep{cho2014learning}, and transformers \citep{vaswani2017attention} which could serve as feature extractors for a DKL model, often assume regularly sampled data. Thus, GP-based solutions require both rich features and computationally tractable representations.

Combining pre-trained representations with a DKL algorithm provides effective dimensionality reduction. We propose the following pipeline. Clinical-BERT \citep{huang2019clinicalbert} first embeds linguistic features from EHRs, such as diagnoses and medications. Features are then aggregated into $\reals{d}$. Observations are mapped into sequences $\randvecx_{1:t} \in \reals{t \times d}$ for lengths $t = 1, \dots, T$ for a population member with $T$ total observations. An encoder $enc_{\theta}: \reals{t \times d} \to \reals{t \times k}$ with parameters $\theta$ and decoder $dec_{\theta^{\prime}}: \reals{t \times k} \to \reals{m}$ with parameters $\theta^{\prime}$ compress features to a latent space with dimension $m \ll d$ at time $t$. The resulting feature extractor $h_{fe}: \reals{t \times d} \to \reals{m}$, is then used in a kernel $k_{fe}$ for a backend GP with a constant mean $m(\randvecx)$:

\begin{align}
    f(\randvecx_{1:t}) & \sim \gp{m(\randvecx_{1:t})}{k_{fe}(\randvecx_{1:t}, \randvecx_{1:t}^{\prime})} \\
    k_{fe}(\randvecx_{1:t}, \randvecx_{1:t}^{\prime}) & = \exp\left(-\frac{\normltwo{h_{fe}(\randvecx_{1:t})}{h_{fe}(\randvecx_{1:t}^{\prime})}}{2l}\right) \\
    h_{fe}(\randvecx_{1:t}) & = dec_{\theta^{\prime}}(enc_{\theta}(\randvecx_{1:t}))
\end{align}

\noindent We require only temporal ordering: $t > t^{\prime}$ implies $t$ occurs after $t^{\prime}$. With fixed embeddings, we optimize the feature extractor parameters $\theta \cup \theta^{\prime}$ via variational inference \citet{wilson2016deep, al2017learning}.

\begin{figure}[tbp]
    \centering
    \subfigure[Deep Kernel Learning Pipeline]{
        \includegraphics[height=7cm, width=0.45\textwidth, trim={0mm, 2cm, 0mm, 3cm}, clip]{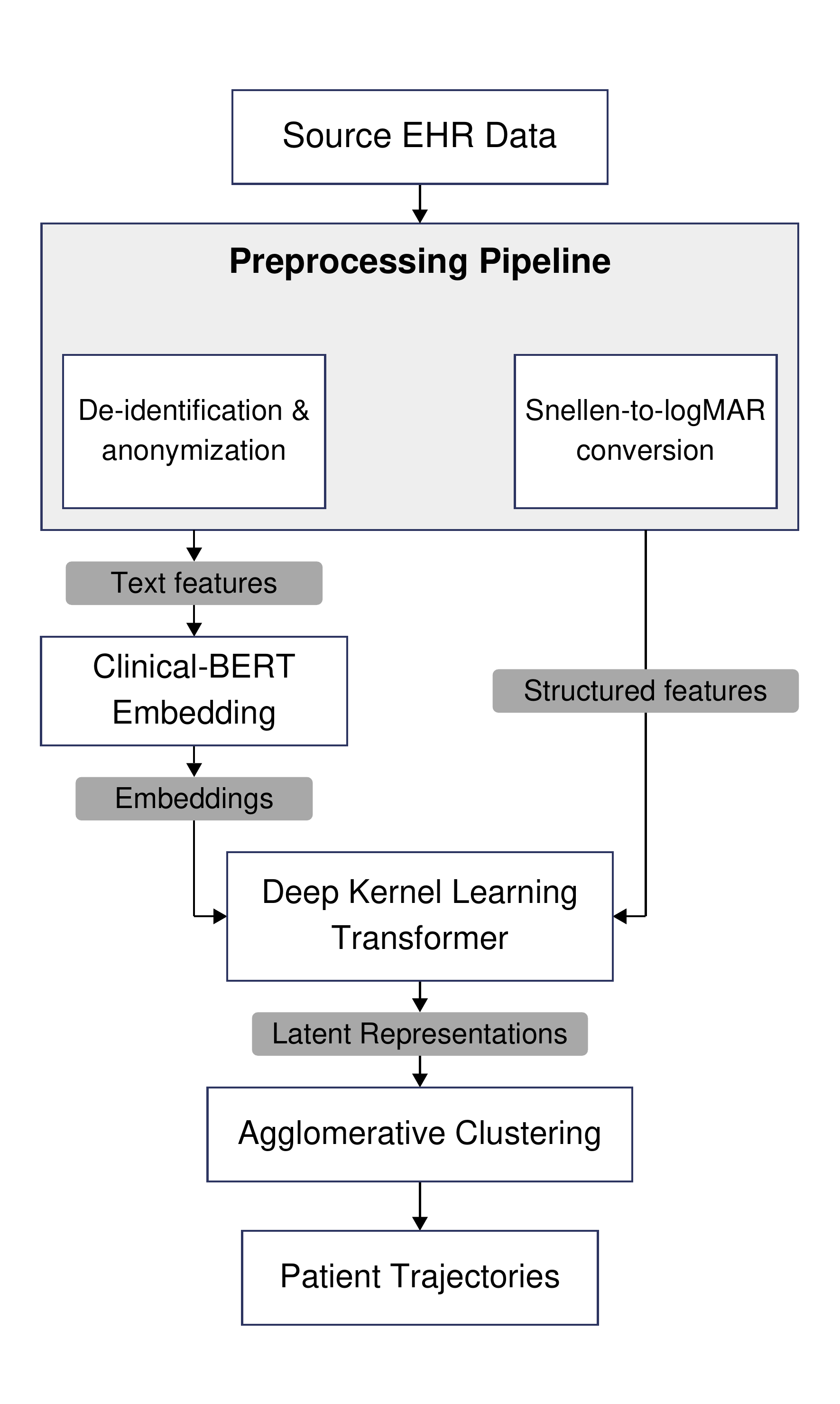}
        \label{fig:dkl-pipeline}
    }%
    \hfill
    \subfigure[Clustering Pipeline for Latent Space Representations]{
        \resizebox{0.45\textwidth}{7cm}{\input{Figures/dkl_clustering_flowchart}}
        \label{fig:clustering_pipeline}
    }
    \caption{\textbf{Deep Kernel Learning Transformer Pipeline for Disease Trajectory Maps}. (a) The transformer architecture processes multimodal EHR data through Clinical-BERT embeddings and structured feature extraction. (b) Agglomerative clustering with ward linkage is applied to latent representations to identify distinct patient trajectories, enabling population-level analysis of clinical patterns and outcomes. Model selection was performed across $c \in \{2,3,4,5\}$ clusters, with $c = 3$ providing optimal separation.}
    \label{fig:dkl-combined}
\end{figure}

While the GP layer in the DKL model naturally handles irregularly sampled time-series data, temporal information still needs to be processed through the DKL encoder. One such strategy is to use recurrent architectures such LSTM and GRU \citep{al2017learning}; another strategy is to utilize transformers \citep{lyu2023efficient}. For time-series data, we often impute missing entries to improve the regularity of the prediction, but due to the GP layer, we can avoid that strategy and instead adopt explicit temporal features by cyclical encoding \citep{fisher1995statistical}: 

\begin{equation}
    f_{\sin}(t) = \sin \left( \frac{2\pi t}{P} \right)
\end{equation}
    
\begin{equation}
    f_{\cos}(t) = \cos \left( \frac{2\pi t}{P} \right)
\end{equation}

\noindent where $P$ is the period (31 for days, 12 for months, and 10 for years). This was performed so that we could better compare DKL algorithms with their standard counterparts, since normal architectures struggle with irregular time-series data.

%% file: Figures/dkl_clustering_flowchart.tex
\begin{tikzpicture}[node distance=1.0cm]

\node[font=\small\bfseries, align=center] (title) {Deep Kernel Learning Pipeline\\for Patient Trajectory Clustering};

\node[data, below=0.6cm of title] (dataset) {Full Dataset};
\node[process, below of=dataset] (sample) {Random Sampling $N = 3821$ patients};
\node[process, below of=sample] (records) {Patient Sequences};
\node[process, below of=records] (model) {Deep Kernel Learning Model};

\node[data, below=1.0cm of model] (means) {Posterior $\mu$};
\node[data, left=1cm of means] (latents) {Latent $Z$};
\node[data, right=1cm of means] (variances) {Posterior $\sigma^{2}$};

\node[process, below=1.0cm of means] (clustering) {Agglomerative Ward Clustering};

\draw[arrow] (dataset) -- (sample);
\draw[arrow] (sample) -- (records);
\draw[arrow] (records) -- (model);

\coordinate[below=0.5cm of model] (fork);
\draw[-] (model) -- (fork);
\draw[arrow] (fork) to[out=175, in=90] (latents.north);
\draw[arrow] (fork) -- (means.north);
\draw[arrow] (fork) to[out=05, in=90] (variances.north);

\coordinate[above=0.5cm of clustering] (merge);
\draw[-] (latents.south) to[out=-90, in=160] (merge);
\draw[-] (means.south) -- (merge);
\draw[-] (variances.south) to[out=-90, in=20] (merge);
\draw[arrow] (merge) -- (clustering);

\end{tikzpicture}

%% file: Sections/experiments.tex
\begin{table}[tbp]
\centering
\begin{tabular}{lcccc}
\toprule
\textbf{Model} & \textbf{MSE} & \textbf{MAE} & $\mathbf{R^{2}}$ & $\mathbf{+/-}$ \textbf{0.1} \textbf{logMAR} \\
\midrule
MLE RNN & 0.1166 & 0.1795 & 0.2594 & 53.8397\\
DKL RNN & 0.1075 & 0.1721 & 0.3172 & 56.9478\\
\midrule
MLE LSTM & 0.1156 & 0.1718 & 0.2659 & 57.9248\\
DKL LSTM & 0.1184 & 0.1712 & 0.2477 & \textbf{59.6589}\\
\midrule
MLE GRU & 0.1106 & 0.1708 & 0.2973 & 56.41\\
DKL GRU & 0.1113 & 0.1748 & 0.2934 & 53.7741\\
\midrule
MLE Transformer & 0.1068 & 0.171 & 0.3217 & 53.3605\\
DKL Transformer & \textbf{0.1061} & \textbf{0.1703} & \textbf{0.3262} & 53.0564\\
\bottomrule
 MLE GRU-D & 0.1246 & 0.1765 & 0.2084 &	56.8392\\
 MLE GRU ODE & 0.1166& 0.1888& 0.2599& 43.8709\\
 MLE RNN CDE & 0.1246& 0.2017& 0.2109& 
36.2228\\
\bottomrule
\end{tabular}
\caption{Average final hold-out test set performance comparison of DKL algorithms trained via variational inference and baseline models using standard maximum likelihood training with stochastic gradient descent across different neural network architectures. All models were run across 10 random seeds and the final test set scores averaged. DKL models demonstrate competitive performance on glaucoma patient data, with transformer-based architectures achieving the lowest mean squared error, mean absolute error, and highest coefficient of determination. The last column is the percentage of predictions within 0.1 logMAR, which indicates clinical accuracy, where LSTM architectures show superior precision in visual acuity prediction for glaucoma patients.}
\label{tab:compact_results}
\end{table}

\section{Experiments}

Our methods were tested on electronic health records (EHRs) from the Sight Outcomes Research Collaborative  (SOURCE) \citep{bommakanti2020application, wang2023prediction}, a multi-institutional EHR initiative led by the University of Michigan that supports the investigation of ophthalmic outcomes. We restricted our analysis to patients receiving care within the [health system redacted for anonymous review] between January 2016 and December 2023. We performed data preprocessing and ran experiments on a variety of architectures and used the best architectures for clustering done via functional data analysis.

\subsection{Data}

The dataset comprises 67,691 patients with 402,552 single-day clinical encounters (henceforth, ``records'') across 18 fields. Structured features include demographics (age, sex, race, ethnicity), temporal data (encounter date, birth date), and clinical metadata (living status, patient ID). Visit features capture specialty, visit type, and free-text reason. Clinical features encompass diagnoses, procedures, medications, lab results, and surgical history. The outcome variable, ACUITY\_SCORE, represents best-corrected visual acuity converted from Snellen to logMAR format. Lower logMAR values indicate better visual acuity; higher logMAR values indicate poorer acuity. For each eye, the best-recorded acuity was defined as the minimum logMAR (corresponding to the smallest Snellen denominator), and the worst acuity as the maximum logMAR (largest Snellen denominator).

\begin{figure}[tbp]
    \centering
    \subfigure[Latent Space Representation of Patient Trajectories]{
        \includegraphics[height=7cm, width=0.45\textwidth]{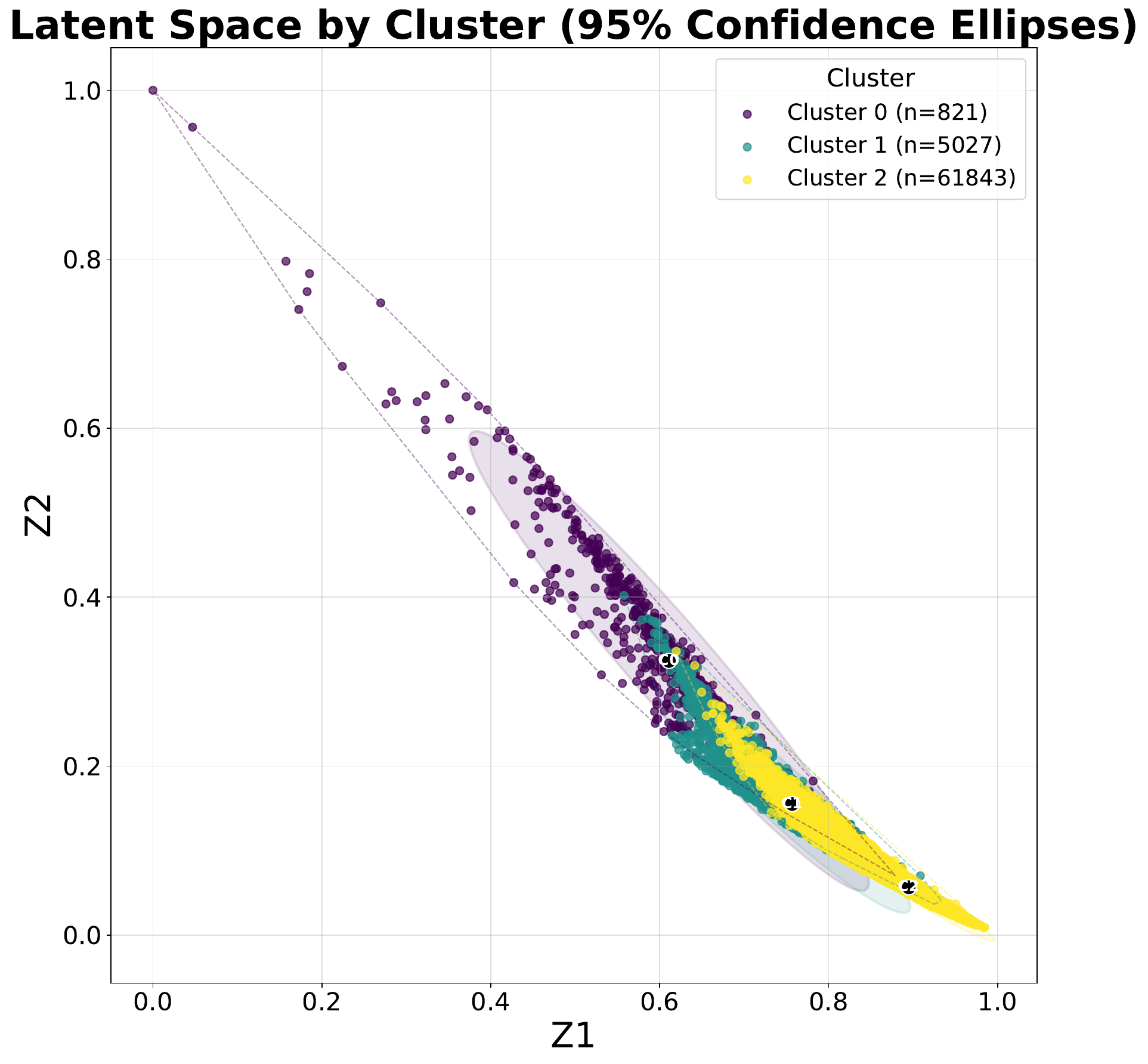}
        \label{fig:latent-space-by-cluster}
    }
    \hfill
    \subfigure[UMAP Projection]{
        \includegraphics[height=7cm, width=0.45\textwidth]{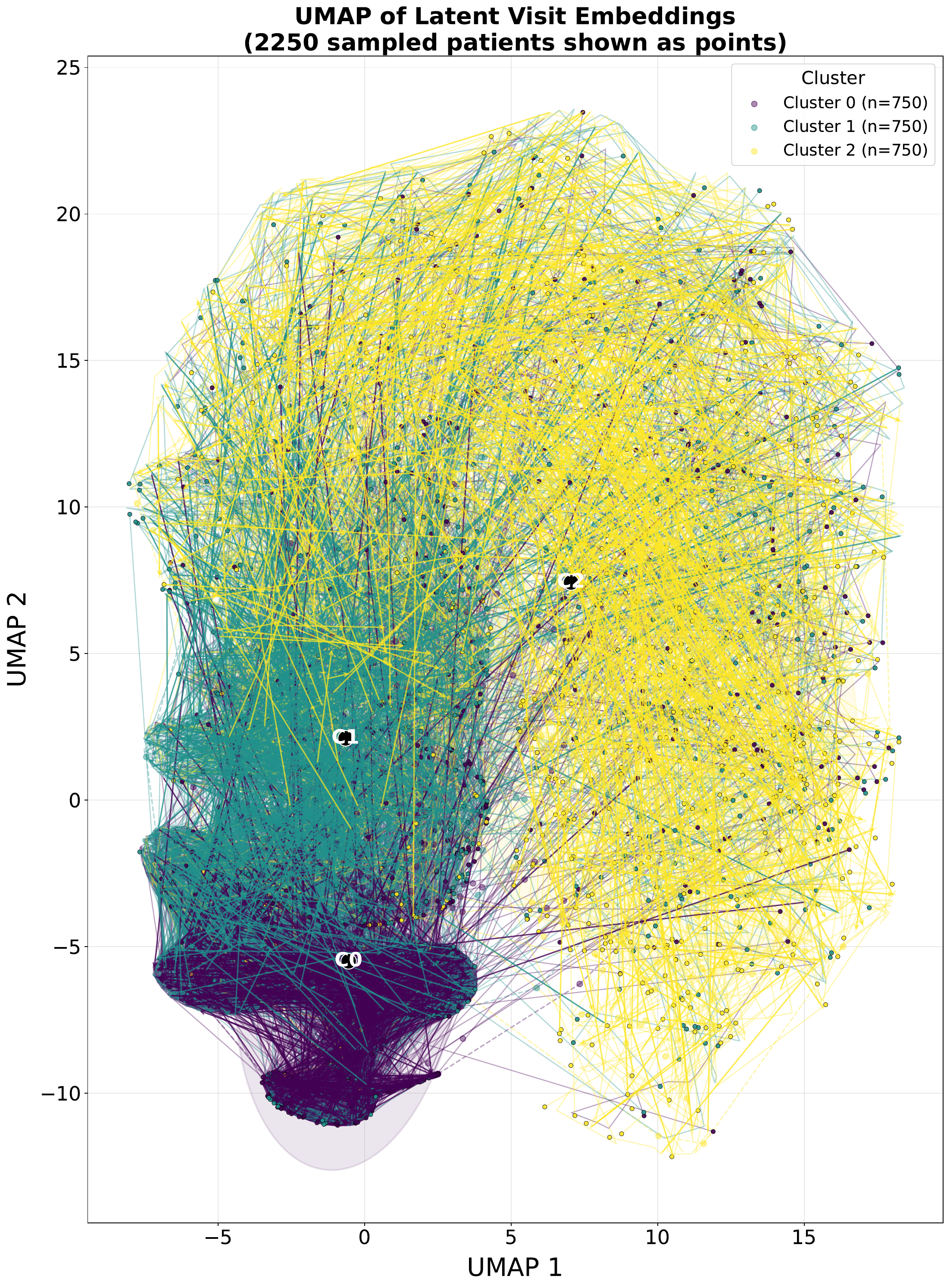}
        \label{fig:umap-latent-space}
    }
    \caption{\textbf{Latent space visualizations of patient trajectory clustering}. (a) Direct latent space visualization demonstrates nonlinear progression patterns along a curved manifold structure with clear separation between three disease progression archetypes. (b) UMAP dimensionality reduction confirms three unique clinical trajectories rather than a continuous spectrum of disease progression.}
    \label{fig:latent-space-combined}
\end{figure}

Data was anonymized through the removal of all patient identifiers and personally identifiable information before processing and other identifiers were randomized to protect patient privacy. Preprocessing handled irregular sampling via duplicate removal, temporal alignment on encounter datetime, and standardized missing values (-1 for numeric, empty strings for text). Visual acuity conversion handled standard fractions (e.g., 20/40) and special cases (NLP, LP, HM, CF). ACUITY\_SCORE used the minimum logMAR across 18 measurement types. 

Clinical text was embedded using Bio\_ClinicalBERT with mean pooling, yielding 768-dimensional representations (batch size: 32, max tokens: 512). Seven text fields were embedded: specialty, visit type, reason, procedures, diagnoses, medications, and lab results. Lab results combined names with values before embedding; medications and procedures were concatenated. This approach preserved clinical semantics while enabling both GP uncertainty quantification and neural feature extraction.

\subsection{Methods}

\begin{figure}[tbp]
    \centering
    \subfigure[Visual Acuity Trajectories]{
        \includegraphics[width=0.45\textwidth]{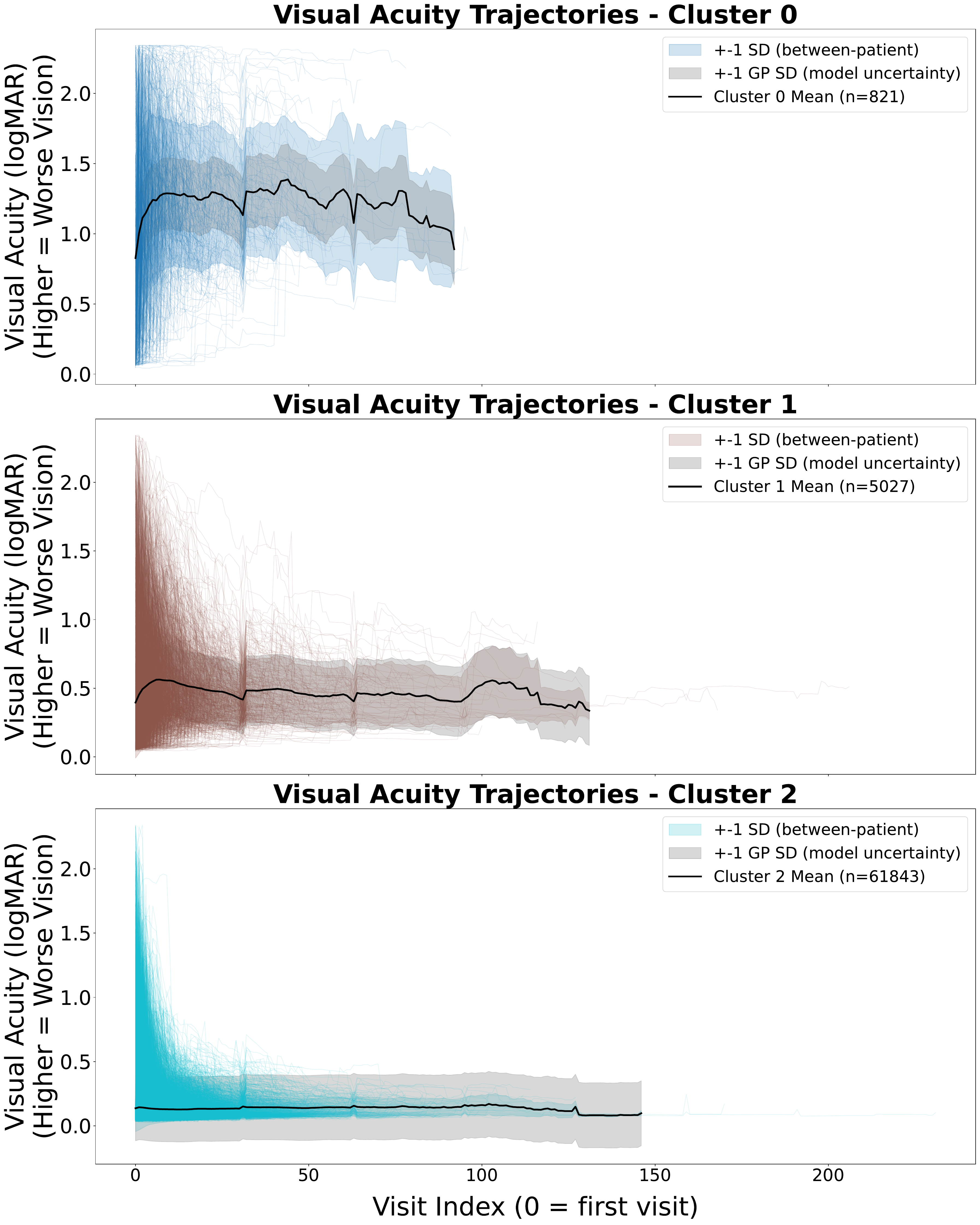}
        \label{fig:visual-acuity-scores}
    }
    \hfill
    \subfigure[SHAP Value Analysis]{
        \includegraphics[width=0.45\textwidth]{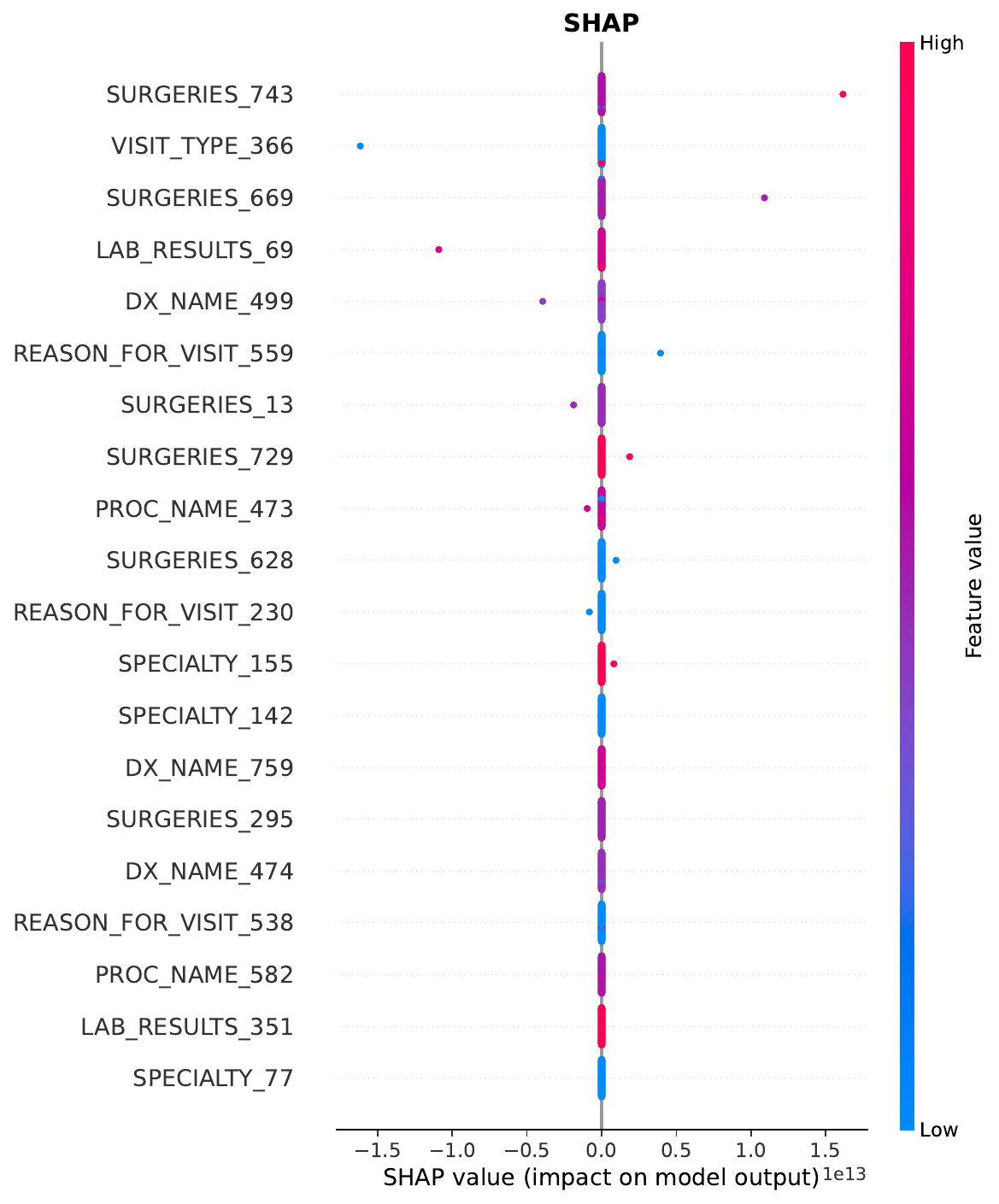}
        \label{fig:shap}
    }
    \caption{\textbf{Clinical trajectory analysis and model interpretability}. (a) Posterior predictive mean trajectories from DKL transformer demonstrate three distinct patient archetypes with characteristic visual acuity patterns. Mean and standard deviation are conditioned on entire cluster. (b) SHAP analysis reveals surgery-related features and specialty codes as primary drivers of model predictions.}
    \label{fig:trajectories-and-shap}
\end{figure}

We compared recurrent neural network (RNN), gated recurrent unit (GRU), long short-term memory (LSTM), and transformer encoder architectures as feature extractors. Baseline models used identical architectures but replaced the GP layer with linear regression trained via maximum likelihood estimation (MLE). Additionally, we trained a GRU-D \citep{che2018recurrent}, GRU ODE \citep{chen2018neural, de2019gru}, and RNN CDE \citep{kidger2020neural} as baselines. DKL models were trained by maximizing the ELBO using stochastic variational inference. We used a multivariate normal variational distribution with Cholesky-decomposed covariances. Our likelihood was a univariate normal distribution.

Both model types used cyclical encoding for temporal features and standardized continuous variables. Cyclical encoding provided temporal inductive bias for baselines while enabling GP benefit assessment. Age and embedding features were z-normalized.

Hyperparameters were selected via grid search on 100,000 samples.  See Appendix A for search details and configurations. We evaluated 1,306 hyperparameter combinations. Models were selected based on the best validation set mean-squared error (MSE).

Models were evaluated using a 70/20/10 train/validation/test split. Performance was assessed with MSE, MAE, \(R^2\), and clinical accuracy, defined as the proportion of predictions within \(\pm 0.1\) logMAR of the observed value. For the DKL models, probabilistic performance was additionally evaluated using 95\% credible interval coverage and the continuous ranked probability score (CRPS). All experiments were repeated across 10 random seeds, and we report the mean and standard deviation across seeds. The MLE and DKL RNN, LSTM, GRU, and Transformer models, as well as the MLE GRU-D model, were trained for 200 epochs per seed. Due to longer training times and faster convergence, the GRU-ODE and RNN-CDE models were trained for 25 epochs per seed.

The full pipeline, from start-to-finish can be seen in figure \ref{fig:dkl-pipeline}.

For clustering, we relied upon functional data analysis and agglomerative ward clustering. The clustering pipeline can be seen in figure \ref{fig:clustering_pipeline}. After training the model, we collected the model's feature extractor latent representation as well as the posterior predictive mean and variance for each record in the patient's sequence of records. We clustered all of the patient trajectory representations using agglomerative ward clustering with the $L^{2}$ norm for distance measurements between trajectories, and we evaluated $c = \{2,3,4,5\}$ clusters, with $c = 3$ clusters ultimately selected due to best clustering metrics and clinical interpretation (see appendix C for more details).

\subsection{Results}

DKL models with transformer extractor outperformed baselines on most metrics (see table \ref{tab:compact_results}). The DKL LSTM outperformed all other methods on clinical accuracy (59.66\%). Compared to MLE, we see that the transformer and RNN feature extractors outperform their MLE counterparts across 10 random seeds, with the RNN improvements being substantial. We found the DKL RNN, GRU, and transformer along with MLE GRU and transformer with encoded temporal features outperformed GRU-D, GRU ODE, and RNN CDE baselines substantially---despite these baselines specialization for irregular time series. See Appendix D for further analyses.
We used the best-performing DKL transformer for trajectory analysis. Figure \ref{fig:latent-space-by-cluster} shows three distinct patient clusters. The latent space reveals three patient archetypes: worsening, stable, and intermediate trajectories (see Appendix B for 3D visualizations). UMAP visualization (Figure \ref{fig:umap-latent-space}) confirms cluster separation in latent space.

Figure \ref{fig:visual-acuity-scores} shows cluster-specific posterior means and uncertainties over time.  Cluster 1 shows progressive worsening before stabilization and exhibits the highest uncertainty. Higher variance reflects heterogeneous progression patterns within this cluster.

To evaluate clinical markers, we performed SHAP analysis \citep{lundberg2017unified} to assess feature relevance for model predictions on the DKL transformer. We found that surgery was most predictive in figure \ref{fig:shap}. To further validate model performance, we experimented with ablations on these features. Table 2 shows our results compared to non-ablated model average performance. We found the largest drop in performance when ablating medicine and surgery names.

%% file: Sections/discussion.tex
\section{Discussion}

The central challenge of applying machine learning to real-world clinical data is not merely managing sparsity or irregularity, but embracing the inherent variability of chronic disease. Patient trajectories are shaped by complex, often unobserved factors, leading to high inter- and intra-patient variance that is sometimes irreducible, even with complete data. Here, we present a probabilistic deep learning framework that, rather than being confounded by this ``messy'' data, leverages it to uncover clinically significant insights. We demonstrate that our approach moves beyond static risk assessment by learning to decouple a patient's current disease state from their future risk of progression.

\paragraph{Handling Irreducible Uncertainty in Sequential Medical Data.}
Our deep kernel learning architecture addresses a fundamental gap in medical ML: most approaches to disease progression either assume regular sampling (violated in real-world EHRs) or provide point estimates without uncertainty quantification (inadequate for clinical decision-making) \citep{shukla2019interpolation}. By combining a transformer-based feature extractor with a Gaussian Process backend, we achieve three critical capabilities: (1) representation learning from high-dimensional, multimodal data including unstructured text, (2) natural handling of irregular temporal sampling without imputation, and (3) calibrated uncertainty estimates that distinguish between aleatory uncertainty (inherent disease variability) and epistemic uncertainty (model confidence). This distinction is crucial—high variance in a patient's trajectory may indicate genuinely unpredictable disease behavior rather than model limitations, requiring fundamentally different clinical responses. The architecture's ability to process sequences of varying lengths with missing features while maintaining uncertainty calibration represents a significant advance over existing methods that typically handle these challenges separately or incompletely.

\begin{table}[tbp]
\centering
\begin{tabular}{lc}
\toprule
\textbf{Feature Category Removed} & \textbf{MSE} \\
\midrule
None (Baseline) & \textbf{0.1081} \\
\midrule
MED\_NAME & 0.1131 \\
SURGERIES & 0.1126 \\
VISIT\_TYPE & 0.1119 \\
PROC\_NAME & 0.1117 \\
DX\_NAME & 0.1100 \\
SPECIALTY & 0.1079 \\
REASON\_FOR\_VISIT & 0.1072 \\
AGE & 0.1071 \\
LAB\_RESULTS & 0.1063 \\
\bottomrule
\end{tabular}
\label{tab:ablation}
\caption{\textbf{Ablation study results showing model performance (MSE) when individual feature categories are removed}. Lower MSE indicates better performance. The baseline model (all features) achieves MSE = 0.1081. Removing medication names (MED\_NAME) or surgeries causes the largest performance degradation, indicating these are the most critical feature categories for prediction accuracy. All experiments on fixed random seed (42).}
\end{table}

\paragraph{Decoupling Progression Risk from Disease State.}
Our primary finding validates the model's ability to learn clinically meaningful representations: clustering in the DKL latent space reveals three distinct patient trajectories that decouple current disease severity from progression risk. Counterintuitively, the group with the worst average visual acuity was not the group with the most severe predicted progression. This demonstrates that our model learns temporal dynamics beyond simple trend extrapolation. This learned separation between current severity and future risk recapitulates a central insight from decades of prospective glaucoma clinical trials. The Canadian Glaucoma Study identified higher baseline IOP, older age, and disc hemorrhages as predictors of open-angle glaucoma progression---but not baseline visual field severity alone \citep{chauhan2008canadian}. Similarly, the Early Manifest Glaucoma Trial (EMGT) found that baseline IOP, pseudoexfoliation, bilateral disease, and age predicted progression, while 45\% of treated patients still progressed despite IOP lowering, underscoring that severity at presentation does not determine trajectory \citep{Leske2003, Leske2007}. The Ocular Hypertension Treatment Study (OHTS) demonstrated that among patients with no baseline glaucomatous damage, a constellation of risk factors — including IOP, central corneal thickness, vertical cup-to-disc ratio, and pattern standard deviation on visual field testing — predicted conversion to glaucoma, establishing that risk is multifactorial and not solely determined by the current disease state \citep{gordon2002ocular}. A large retrospective analysis combining data from AGIS and a clinical cohort found that after adjusting for IOP, baseline visual field severity did not independently predict the rate of visual field change, further supporting the dissociation between where a patient is and where they are going \citep{DeMoraes2011}. 

Collectively, these trials demonstrate that baseline disease severity is a poor proxy for future progression risk — a principle well understood by glaucoma specialists, but difficult to operationalize in routine clinical practice, particularly at the population level. Our framework provides a computational approach to this clinical problem. In the current research, we identify: (1) a ``Stable-Good Vision'' group with low vision loss and tight confidence intervals consistent with well-controlled or non-progressive disease, (2) a ``High-Progression Risk'' group with moderate vision but high trajectory variance, consistent with volatile disease courses requiring intensive monitoring and potential escalation of treatment, and (3) a ``Stable-Poor Vision'' group with chronically poor but non-worsening vision, potentially representing patients with end-stage but quiescent disease, or patients whose poor acuity reflects non-glaucomatous pathology such as visually significant cataract or macular disease. That an unsupervised model — without access to IOP measurements, visual fields, or OCT data — recovers trajectory subtypes consistent with the progression risk factors identified by landmark clinical trials provides external validation that our learned representations capture genuine disease heterogeneity, rather than statistical artifacts. Additionally, it suggests that the multimodal EHR features processed through Clinical-BERT embeddings encode clinically meaningful information about disease dynamics that is partially redundant with, and may complement, the structured clinical measurements used in traditional risk calculators.

\paragraph{Quantifying and Leveraging Trajectory Uncertainty.}
The ``High-Progression Risk'' cluster exemplifies our framework's novel contribution to uncertainty-aware ML. These patients exhibit high trajectory variance—not due to sparse data or model uncertainty, but reflecting genuinely volatile disease courses. Traditional ML approaches would either smooth over this variability (losing critical information) or produce unstable predictions (limiting clinical utility). Our GP backend quantifies this uncertainty, enabling risk-aware decision support: patients with wide posterior credible intervals trigger more frequent monitoring, while those with narrow bands support extended follow-up. This moves beyond binary classification to provide nuanced, uncertainty-aware recommendations that align with clinical decision-making under uncertainty. The model achieves 53.06\% accuracy within 0.1 logMAR (approximately one line on the eye chart), exceeding typical progression prediction tools that achieve AUCs of 0.65-0.75 \citep{yohannan2017evidence}, while additionally providing calibrated uncertainty estimates. Notably, the clinical significance of a ±0.1 logMAR prediction error varies across the acuity spectrum. In the context of glaucoma trajectory prediction — where the goal is to identify patients trending toward visual impairment (logMAR $\ge$0.5) or legal blindness (logMAR $\ge$1.0) — this level of accuracy is sufficient for clinically meaningful risk stratification, even if insufficient for precise acuity prediction.

\paragraph{Digital Biomarkers from Multimodal Temporal Learning.}
The learned patient trajectories represent novel digital biomarkers that integrate the full complexity of longitudinal clinical data \citep{purushotham2018benchmarking}. Unlike traditional risk factors (central corneal thickness, cup-to-disc ratio, baseline IOP \citep{gordon2002ocular}), our approach leverages Clinical-BERT embeddings to incorporate unstructured clinical notes—capturing subtle observations about disease stability, treatment response, and psychosocial factors typically unavailable to structured models. The transformer architecture learns temporal dependencies across irregular visit sequences, while the DKL framework maps these high-dimensional histories to interpretable low-dimensional trajectories. This demonstrates that modern NLP and sequence modeling techniques can extract clinically meaningful signals from the ``messy'' reality of EHR data, where crucial information is often buried in narrative text rather than structured fields.

\subsection{Generalizable Framework for Chronic Disease Trajectories.}
While demonstrated on glaucoma, our framework addresses challenges universal to chronic disease modeling. The combination of irregular sampling, high-dimensional multimodal data, and irreducible patient-level variability characterizes conditions from diabetes to heart failure to rheumatoid arthritis. Our architecture's key innovations provide a template for chronic disease analytics. The success in identifying clinically meaningful subtypes without supervision suggests that similar latent structure may exist across chronic conditions, waiting to be discovered by appropriate ML methods. Future work could explore transfer learning across diseases, potentially sharing learned representations about healthcare utilization patterns, treatment response, and disease progression dynamics.

Given glaucoma's disproportionate impact on African American and Latino populations (prevalence 3-4 times higher than Caucasians \citep{tielsch1991racial}), our work explicitly addresses algorithmic fairness—a critical requirement for clinical ML deployment. Our framework enables systematic fairness audits by: (1) including demographic variables as model inputs, (2) analyzing performance metrics stratified by protected attributes, (3) examining cluster membership distributions across demographic groups, and (4) quantifying whether trajectory uncertainty correlates with social determinants of health. This transparency is essential for responsible AI in healthcare, moving beyond post-hoc bias detection to proactive fairness-aware modeling. The probabilistic nature of our approach allows uncertainty to increase in underrepresented populations, honestly reflecting data limitations rather than producing falsely confident predictions. Preliminary analysis of cluster demographics is ongoing. Given that the SOURCE dataset includes patients from the University of Virginia health system, which serves a diverse catchment area including rural Appalachian communities and urban populations, characterizing whether the identified trajectory subtypes differ across demographic groups is essential for ensuring equitable clinical deployment.

Our current limitations highlight important directions for medical ML research. The lack of explicit inter-eye correlation modeling (common in bilateral diseases) motivates development of multi-output GP architectures for correlated outcomes. Reliance on visual acuity rather than gold-standard endpoints (visual fields, OCT measurements) demonstrates the need for methods that can align heterogeneous clinical endpoints with different sampling frequencies and measurement noise characteristics. An important current limitation of this study is the use of visual acuity as the primary outcome measure for glaucoma trajectory modeling. Visual acuity is a late and non-specific indicator of glaucomatous damage---patients can lose substantial peripheral vision and retinal nerve fiber layer thickness while maintaining 20/20 central acuity. VA loss in glaucoma typically occurs only with advanced disease involving central visual field damage, or through comorbid conditions such as retinal disease or visually significant cataract. Consequently, the trajectories identified here likely capture a composite of glaucoma progression, cataract development, and other causes of central vision loss rather than glaucoma-specific progression alone. Gold-standard glaucoma progression measures — including visual field mean deviation (MD) trends, OCT retinal nerve fiber layer (RNFL) thinning rates, and intraocular pressure (IOP) trajectories---were not available for this current analysis, but would substantially improve the clinical specificity of identified patient subtypes. Future iterations of the present research incorporating these modalities will enable earlier detection of progression, as structural (OCT) and functional (VF) changes precede VA loss by years in most glaucoma patients. The inability to distinguish glaucoma-driven visual acuity loss from cataract-driven visual acuity loss represents a significant confounding factor in trajectory interpretation. Cataract is highly prevalent in the age demographic most commonly affected by glaucoma, and cataract surgery (often combined with glaucoma procedures) can dramatically improve VA. Patients in the 'volatile' cluster may potentially include individuals whose VA fluctuations reflect cataract progression followed by surgical improvement, rather than glaucomatous instability. Additionally, the strong SHAP signal from surgery-related features (Figure 3b) likely captures both glaucoma-specific interventions and combined cataract-glaucoma procedures, which have fundamentally different clinical implications. Future work should ideally stratify patients by cataract status and separate combined cataract-glaucoma procedures from glaucoma-only surgeries to disambiguate these trajectories. Furthermore, the dominance of surgery-related features in the current SHAP analysis warrants careful causal interpretation. In observational EHR data, glaucoma surgery is a treatment response typically triggered by disease progression or failure of medical management, creating an inherent potential confounder: surgery correlates with poor outcomes partly because it is performed on patients already doing poorly. Though this paradigm is being challenged by more proactive "interventional glaucoma" surgical management paradigms emphasizing early or proactive intervention with (usually) lower-risk minimally invasive glaucoma surgery (MIGS), the complex relationship between decision for surgery and poor outcome is analogous to 'confounding by indication' in pharmacoepidemiology. The model likely learns surgery as a strong proxy for disease severity and prior progression rather than as a causal driver of outcomes. Disentangling this relationship — and ultimately modeling counterfactual trajectories (e.g., 'what if surgery had been performed earlier?' or 'what if MIGS were tried rather than a more invasive surgery such as trabeculectomy or glaucoma tube shunt procedure?') — is a key direction for developing actionable clinical decision support from this framework. The challenge of distinguishing disease-driven from opportunistic interventions (e.g., glaucoma-only vs. combined cataract-glaucoma surgery) represents a broader causal inference problem in observational health data. These limitations are not merely clinical details but fundamental ML challenges in learning from observational data with complex dependencies and confounding \citep{trottet2023modeling}.

%% file: Sections/appendix_a.tex
\section{}

Our feature extractor architectures consisted of recurrent neural networks (RNNs), gated recurrent units (GRUs), long short-term memory (LSTM), and transformer encoders. Our grid search initially started with transformer encoders for prototyping and we expanded to the other architectures. After we found that the GRU and transformer encoder feature extractors had the best performance, per validation mean squared error, we expanded the parameter configuration searches for those architectures to considerably more than the RNN and LSTM models. The total number of parameter configurations search was $1306$ and the different configurations are given in the tables. Optimal parameters for each architecture are in \textbf{bold}:

\begin{table}[h!]
\centering
\caption{Optimal Hyperparameters (bold) for All Feature Extractor Architectures}
\label{tab:all_hyperparams}
\resizebox{\textwidth}{!}{%
\begin{tabular}{@{}lccccccc@{}}
\toprule
\textbf{Hyperparameter} & \textbf{RNN} & \textbf{GRU} & \textbf{LSTM} & \textbf{Transformer} & \textbf{CDE} & \textbf{ODE} & \textbf{GRUD} \\
\midrule
Num.\ Inducing Points  & 64, \textbf{128} & 64, \textbf{128} & 64, \textbf{128} & 64, \textbf{128} & -- & -- & -- \\
Batch Size             & \textbf{32} & \textbf{32} & \textbf{32} & \textbf{32}, 64 & \textbf{32} & \textbf{32} & \textbf{32} \\
Model/Hidden Dim.      & 16, 64, 256, \textbf{512} & 16, 64, 256, \textbf{512} & 16, 64, 256, \textbf{512} & 128, 256, \textbf{512}, 1024 & \textbf{8} & \textbf{256} & \textbf{512} \\
Num.\ Heads            & -- & -- & -- & 4, 8, 16, \textbf{32}, 64, 128 & -- & -- & -- \\
Feedforward Dim.       & -- & -- & -- & 512, 1024, \textbf{2048} & -- & -- & -- \\
Num.\ Layers           & 1, 2, \textbf{3}, 4, 6 & 1, 2, \textbf{3}, 4, 6 & 1, 2, 3, \textbf{4} & 3, 4, \textbf{6}, 8, 12 & -- & -- & -- \\
Decoder Dim.           & 64, \textbf{128}, 256 & 64, \textbf{128}, 256 & 64, \textbf{128}, 256 & 16, 32, 64, 128, \textbf{256} & \textbf{64} & \textbf{32} & \textbf{128} \\
Latent Dim.            & \textbf{2}, 3, 4 & 2, 3, \textbf{4} & 2, \textbf{3}, 4 & \textbf{2}, 3, 4, 5 & \textbf{2} & \textbf{3} & \textbf{4} \\
Learning Rate          & \textbf{5e-5}, 1e-4, 2e-4 & 1e-4, \textbf{2e-4}, 3e-4, 5e-4 & \textbf{5e-5}, 1e-4, 2e-4 & \textbf{1e-4}, 2e-4, 5e-4 & 1e-4, 2e-4, \textbf{3e-4} & \textbf{1e-4}, 2e-4 & \textbf{1e-4}, 2e-4 \\
\bottomrule
\end{tabular}%
}
\end{table}

%% file: Sections/appendix_b.tex
\section{}

We plot the raw and clustered trajectories with respect to the latent representations of a DKL-transformer.

\begin{figure}[h!]
    \centering
    \subfigure[Means]{
        \includegraphics[width=0.45\textwidth]{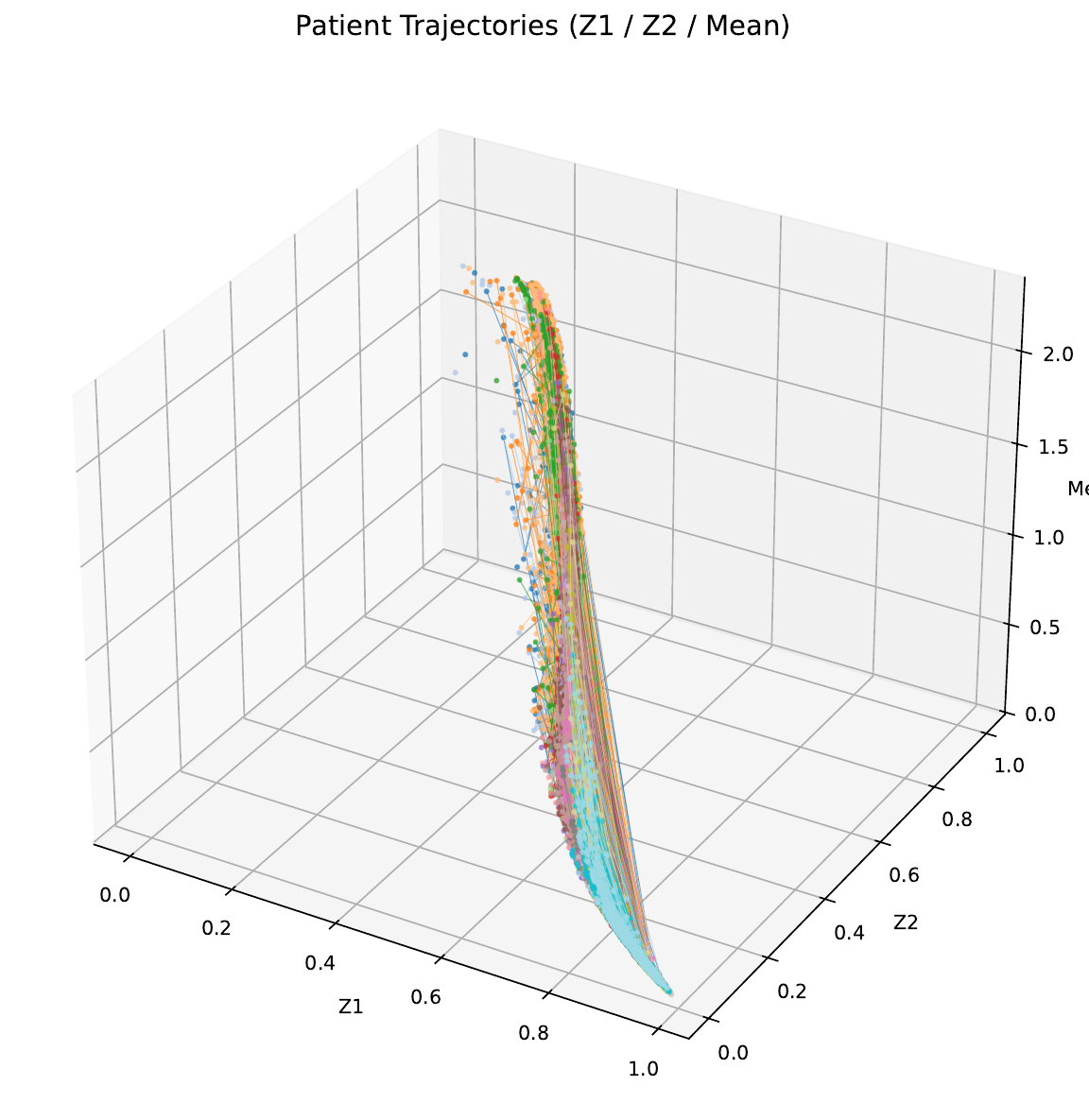}
        \label{fig:raw-means}
    }
    \hfill
    \subfigure[Variances]{
        \includegraphics[width=0.45\textwidth]{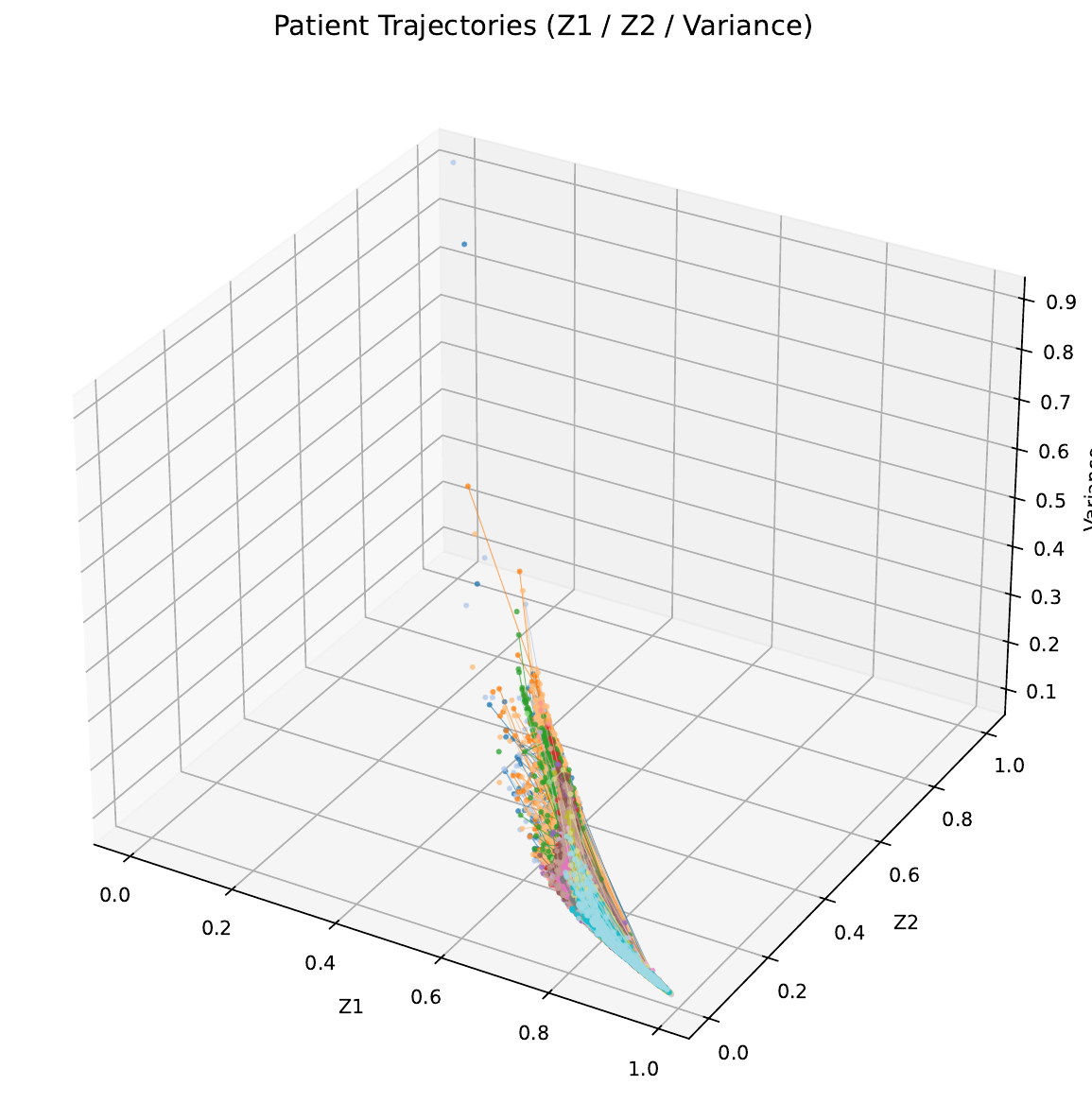}
        \label{fig:raw-variances}
    }    
    \caption{Posterior predictive mean (\ref{fig:raw-means}) and variance (\ref{fig:raw-variances}) trajectories with respect to DKL transformer latent dimensions, renormalized on 3,821 patients. The Z-axis represents mean in logMAR units where higher indicates worse glaucoma.}
    \label{fig:raw}
\end{figure}

\begin{figure}[h!]
    \centering
    \subfigure[Clustered Means]{
        \includegraphics[width=0.45\textwidth]{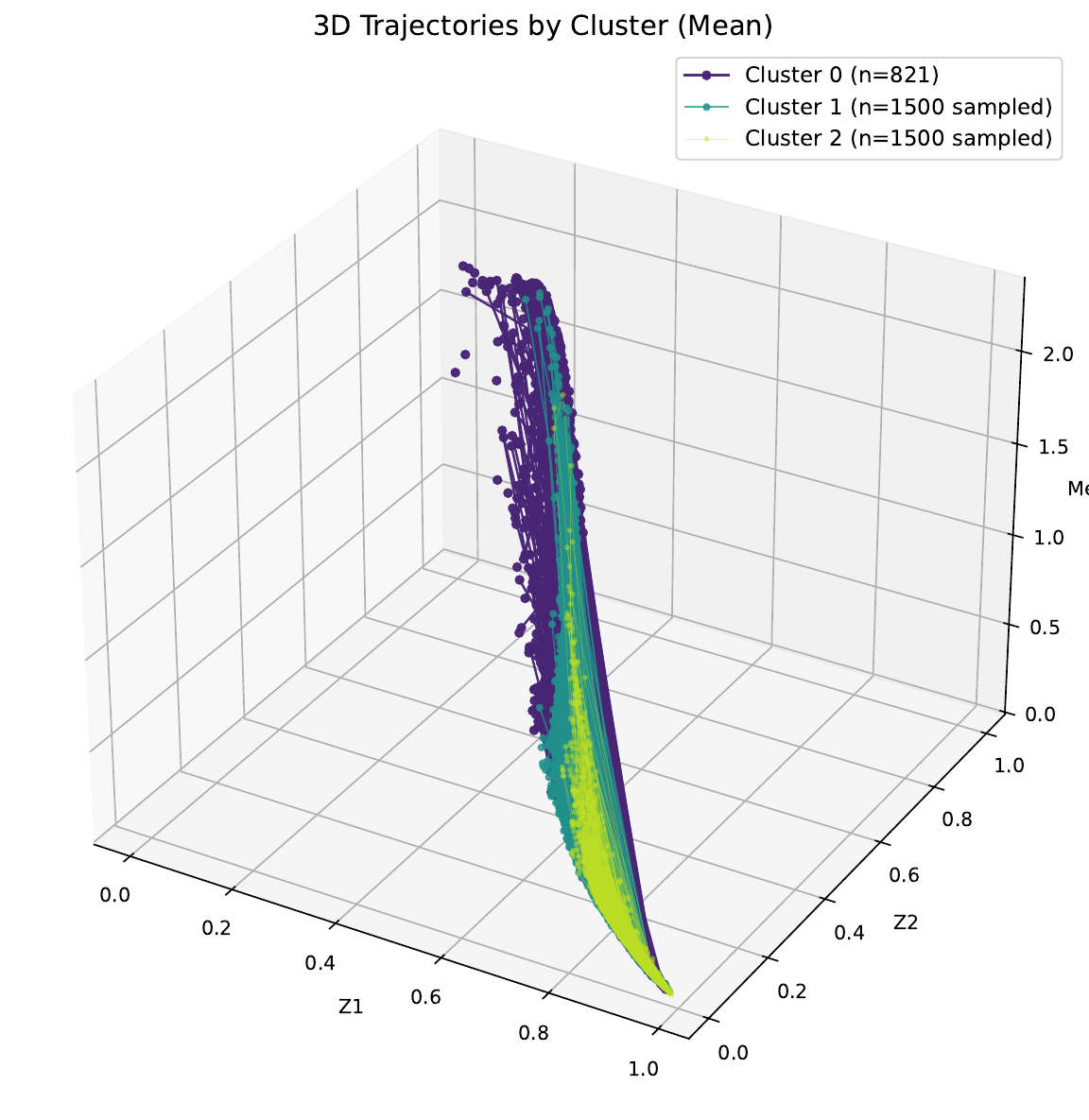}
        \label{fig:clustered-means}
    }
    \hfill
    \subfigure[Clustered Variances]{
        \includegraphics[width=0.45\textwidth]{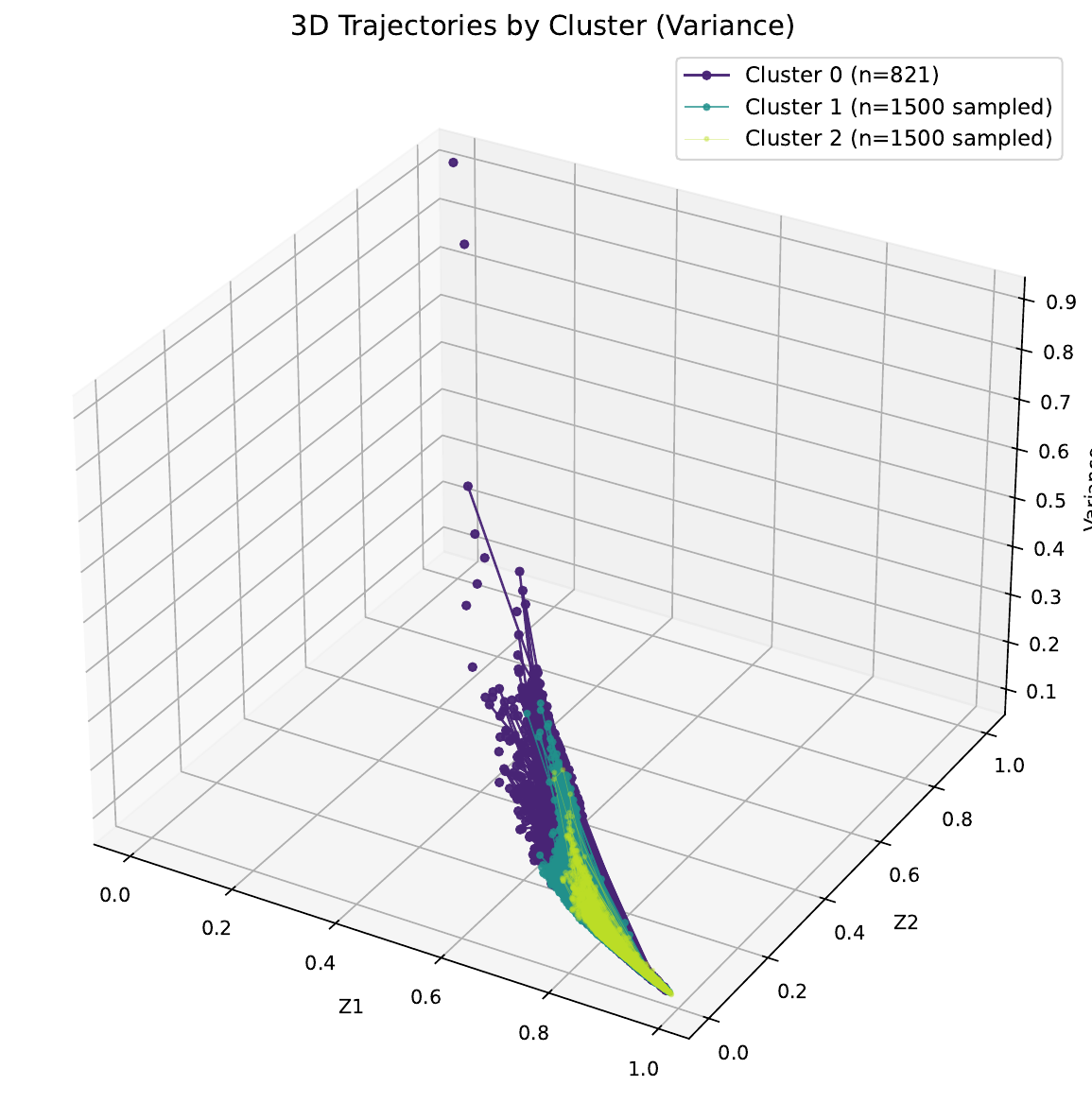}
        \label{fig:clustered-variances}
    }    
    \caption{Posterior predictive mean (\ref{fig:clustered-means}) and variance (\ref{fig:clustered-variances}) trajectories with respect to DKL transformer latent dimensions, renormalized on 3,821 patients. The Z-axis represents mean in logMAR units where higher indicates worse glaucoma. Each trajectory represents a patient and three clusters represent three distinct patient trajectories, with purple indicating worst, blue indicating moderate, and yellow indicating stable.}
    \label{fig:clustered}
\end{figure}

%% file: Sections/appendix_c.tex
\section{}

Here, we provide additional cluster statistics for our experiments. We compared agglomerative ward, agglomerative average linkage, functional k-means, and GMM across $k \in \{2,3,4,5\}$ (Table \ref{tab:clustering_comparison}). While agglomerative average linkage achieved marginally higher Silhouette scores at k=3, it produced severely imbalanced clusters (n=67,033, 508, 150), rendering the solution clinically uninterpretable. Agglomerative ward with k=3 achieved a Silhouette score of 0.743 and Davies-Bouldin score of 0.576, with balanced, clinically meaningful cluster sizes, and was selected as the final configuration.

\pagebreak
\begin{landscape}
\begin{table}[h!]
    \centering
    \caption{Clustering Method Comparison Across $k \in \{2, 3, 4, 5\}$}
    \label{tab:clustering_comparison}
    \begin{tabular}{llccc}
    \toprule
        \textbf{Method} & \textbf{K} & \textbf{Silhouette} $\uparrow$ & \textbf{Davies-Bouldin} $\downarrow$ & \textbf{Cluster Sizes} \\
    \midrule
        Agglomerative Ward & 2 & 0.8983 & 0.3925 & [66870, 821] \\
        Agglomerative Avg  & 2 & 0.9047 & 0.3582 & [67033, 658] \\
        Functional K-Means & 2 & 0.8682 & 0.5184 & [65882, 1809] \\
        GMM                & 2 & 0.6077 & 1.0027 & [55437, 12254] \\
    \midrule
        \textbf{Agglomerative Ward} & \textbf{3} & \textbf{0.7428} & \textbf{0.5756} & \textbf{[61843, 5027, 821]} \\
        Agglomerative Avg  & 3 & 0.8863 & 0.4565 & [67033, 508, 150] \\
        Functional K-Means & 3 & 0.7349 & 0.5795 & [60513, 6474, 704] \\
        GMM                & 3 & 0.4880 & 1.0280 & [52738, 11438, 3515] \\
    \midrule
        Agglomerative Ward & 4 & 0.5366 & 0.6918 & [48155, 13688, 5027, 821] \\
        Functional K-Means & 4 & 0.6380 & 0.6172 & [54369, 10994, 1882, 446] \\
        GMM                & 4 & 0.3682 & 1.3219 & [46946, 11142, 7556, 2047] \\
    \midrule
        Agglomerative Ward & 5 & 0.5369 & 0.6456 & [48155, 13688, 5027, 665, 156] \\
        Functional K-Means & 5 & 0.5353 & 0.6531 & [43890, 18026, 4349, 1103, 323] \\
        GMM                & 5 & 0.2978 & 1.3383 & [37764, 16749, 8761, 3415, 1002] \\
    \bottomrule
    \end{tabular}
    \vspace{0.2cm}
        \begin{tablenotes}
            \small
            \item \textit{Note: Silhouette scores range from $-1$ to $1$; higher is better. Davies-Bouldin scores measure intra-cluster similarity relative to inter-cluster distance; lower is better. Agglomerative average linkage at $k=3$ achieves a higher Silhouette score but produces highly imbalanced clusters (n=67,033, 508, 150), limiting clinical interpretability. Agglomerative ward with $k=3$ (bold) was selected as the final configuration based on its strong cluster quality metrics and clinically meaningful, balanced cluster sizes. Agglomerative average ($k=2,3$) results excluded from $k \geq 4$ as cluster sizes collapse to near-singleton groups.}
        \end{tablenotes}
\end{table}
\end{landscape}

\pagebreak

To assess robustness, we evaluated agglomerative ward stability across 100 random seed initializations, yielding high consistency (NMI: $0.950 \pm 0.120$, ARI: $0.957 \pm 0.103$; Table~\ref{tab:clustering_stability}), confirming the presence of genuine cluster structure rather than algorithmic artifacts.

\begin{table}[h!]
    \centering
    \caption{Clustering Stability Metrics for Transformer + DKL ($k=3$)}
    \label{tab:clustering_stability}
    \begin{tabular}{lcc}
    \toprule
        \textbf{Metric} & \textbf{Mean} & \textbf{Std} \\
    \midrule
        NMI & 0.950 & 0.120 \\
        ARI & 0.957 & 0.103 \\
    \bottomrule
    \end{tabular}
    
    \vspace{0.2cm}
    \small
    \textit{Note: NMI = Normalized Mutual Information; ARI = Adjusted Rand Index. Stability metrics computed across 100 clustering runs with different random initializations for $k=3$ clusters.}
\end{table}

Table~\ref{tab:clustering_metrics} describes the final agglomerative ward ($k=3$) cluster metrics including Silhouette and Calinski-Harabasz scores. 

\begin{table}[h!]
    \centering
    \caption{Clustering Quality Metrics and Overall Dataset Statistics}
    \label{tab:clustering_metrics}
    \begin{tabular}{lr}
    \toprule
        \textbf{Metric} & \textbf{Value} \\
    \midrule
        Total Patients & 67,691\\
        Total Data Points & 402,552\\
        Number of Clusters & 3 \\
    \midrule
        \multicolumn{2}{l}{\textbf{Cluster Distribution:}} \\
        \quad Cluster 0 & 821 patients\\
        \quad Cluster 1 & 5027 patients\\
        \quad Cluster 2 & 61843 patients\\
    \midrule
        \multicolumn{2}{l}{\textbf{Quality Metrics:}} \\
        \quad Silhouette Score & 0.5373\\
        \quad Calinski-Harabasz Score & 23910.874\\
    \bottomrule
    \end{tabular}
    
\end{table}

%% file: Sections/appendix_d.tex
\section{}

To assess model stability, we trained the ODE and CDE architectures for 25 epochs and all other architectures for 200 epochs using 10 random seeds (42, 123, 456, 789, 1024, 2048, 3141, 5926, 8765, and 4321), with each seed producing a different train-validation-test split. Table~\ref{tab:variance_results} reports the mean and standard deviation of test performance across seeds. Table \ref{tab:crps} summarizes probabilistic performance for the DKL models using the continuous ranked probability score (CRPS), reported as the mean and standard deviation across seeds.

\begin{table}[h!]
    \centering
    \caption{CRPS performance across 10 random seeds for DKL models. Lower values indicate better probabilistic calibration and sharpness.}
    \label{tab:crps}
    \begin{tabular}{lcc}
    \toprule
        \textbf{Architecture} & \textbf{CRPS Mean} & \textbf{CRPS SD} \\
    \midrule
        DKL GRU         & 0.0839 & 0.0516 \\
        DKL LSTM        & 0.0936 & 0.0093 \\
        DKL RNN         & 0.0785 & 0.0091 \\
        DKL Transformer & 0.0122 & 0.0291 \\
    \bottomrule
    \end{tabular}
\end{table}

\pagebreak

\begin{landscape}
    \begin{table}[H]
    \centering
    \caption{Experimental Variance and Performance. All models were tested across 10 random seeds. All models were trained to validation loss convergence. We observe that DKL models tended to outperform alternatives.}
    \label{tab:variance_results}
    \begin{tabular}{ccccccc}\toprule
         \textbf{ Architecture}&  \textbf{ Mean MSE}&  \textbf{ MSE SD}&  \textbf{ Mean MAE}&  \textbf{ MAE SD}&  \textbf{ Mean (\%) $\pm$ \textbf{0.1} }& \textbf{SD (\%)  $\pm$ \textbf{0.1} }\\\midrule
         \textbf{ MLE GRU}&  		0.1106&  		0.0036&  		0.1708&  		0.0023&  	56.41& 	1.6192\\
         \textbf{ DKL GRU}&  		0.1113&  		0.0168&  			0.1748&  		0.0197&  		53.7741& 	7.9965\\
          \textbf{ MLE LSTM}&  		0.1156&  		0.0042&  			0.1718&  		0.0037&  	57.9248& 	1.3425\\
         \textbf{ DKL LSTM}&  		0.1184&  		0.0027&  		0.1712&  		0.0018&  		\textbf{59.6589}& 	\textbf{0.4508}\\
         \textbf{ MLE RNN}&  		0.1166&  		0.0062&  		0.1795&  		0.0071&  		53.8397& 	8.3422\\
         \textbf{ DKL RNN}&  		0.1075&  		0.0025&  		0.1721&  		0.0021&  		56.9478& 	1.9967\\
         \textbf{ MLE Transformer}&  		0.1068&  		\textbf{0.0015}&  		0.171&  		\textbf{0.0015}&  	53.3605& 	1.2419\\
         \textbf{DKL Transformer}&  		\textbf{0.1061}&  		0.0026&  			\textbf{0.1703}&  		0.0026&  	53.0564& 		1.3695\\
         \textbf{GRU-D}&  		0.1246&  		0.0047&  		0.1765&  		0.0033&  		56.8392&  	0.8086\\ 
 \textbf{GRU ODE}& 0.1166& 0.0029& 0.1888& 0.0028& 43.8709&2.1627\\
 \textbf{RNN CDE}& 0.1246& 0.0033& 0.2017& 0.0055& 36.2228&2.2172\\ \bottomrule
    \end{tabular}
\end{table}

\end{landscape}

%% file: mlhc.bib
@article{schulam2016disease,
  title={Disease trajectory maps},
  author={Schulam, Peter and Arora, Raman},
  journal={Advances in neural information processing systems},
  volume={29},
  year={2016}
}

@inproceedings{wilson2016deep,
  title={Deep kernel learning},
  author={Wilson, Andrew Gordon and Hu, Zhiting and Salakhutdinov, Ruslan and Xing, Eric P},
  booktitle={Artificial intelligence and statistics},
  pages={370--378},
  year={2016},
  organization={PMLR}
}

@article{vaswani2017attention,
  title={Attention is all you need},
  author={Vaswani, Ashish and Shazeer, Noam and Parmar, Niki and Uszkoreit, Jakob and Jones, Llion and Gomez, Aidan N and Kaiser, {\L}ukasz and Polosukhin, Illia},
  journal={Advances in neural information processing systems},
  volume={30},
  year={2017}
}

@article{trottet2023modeling,
  title={Modeling Complex Disease Trajectories using Deep Generative Models with Semi-Supervised Latent Processes},
  author={Trottet, C{\'e}cile and Sch{\"u}rch, Manuel and Allam, Ahmed and Barua, Imon and Petelytska, Liubov and Distler, Oliver and Hoffmann-Vold, Anna-Maria and Krauthammer, Michael and others},
  journal={arXiv preprint arXiv:2311.08149},
  year={2023}
}

@article{purushotham2018benchmarking,
  title={Benchmarking deep learning models on large healthcare datasets},
  author={Purushotham, Sanjay and Meng, Chuizheng and Che, Zhengping and Liu, Yan},
  journal={Journal of biomedical informatics},
  volume={83},
  pages={112--134},
  year={2018},
  publisher={Elsevier}
}

@article{shukla2019interpolation,
  title={Interpolation-prediction networks for irregularly sampled time series},
  author={Shukla, Satya Narayan and Marlin, Benjamin M},
  journal={arXiv preprint arXiv:1909.07782},
  year={2019}
}

@article{chauhan2008canadian,
  title={Canadian Glaucoma Study: 2. risk factors for the progression of open-angle glaucoma},
  author={Chauhan, Balwantray C and Mikelberg, Frederick S and Balaszi, A Gordon and LeBlanc, Raymond P and Lesk, Mark R and Trope, Graham E and Canadian Glaucoma Study Group and others},
  journal={Archives of ophthalmology},
  volume={126},
  number={8},
  pages={1030--1036},
  year={2008},
  publisher={American Medical Association}
}

@article{yohannan2017evidence,
  title={Evidence-based criteria for assessment of visual field reliability},
  author={Yohannan, Jithin and Wang, Jiangxia and Brown, Jamie and Chauhan, Balwantray C and Boland, Michael V and Friedman, David S and Ramulu, Pradeep Y},
  journal={Ophthalmology},
  volume={124},
  number={11},
  pages={1612--1620},
  year={2017},
  publisher={Elsevier}
}

@article{gordon2002ocular,
  title={The Ocular Hypertension Treatment Study: baseline factors that predict the onset of primary open-angle glaucoma},
  author={Gordon, Mae O and Beiser, Julia A and Brandt, James D and Heuer, Dale K and Higginbotham, Eve J and Johnson, Chris A and Keltner, John L and Miller, J Philip and Parrish, Richard K and Wilson, M Roy and others},
  journal={Archives of ophthalmology},
  volume={120},
  number={6},
  pages={714--720},
  year={2002},
  publisher={American Medical Association}
}

@article{tielsch1991racial,
  title={Racial variations in the prevalence of primary open-angle glaucoma: the Baltimore Eye Survey},
  author={Tielsch, James M and Sommer, Alfred and Katz, Joanne and Royall, Richard M and Quigley, Harry A and Javitt, Jonathan},
  journal={Jama},
  volume={266},
  number={3},
  pages={369--374},
  year={1991},
  publisher={American Medical Association}
}

@article{al2017learning,
  title={Learning scalable deep kernels with recurrent structure},
  author={Al-Shedivat, Maruan and Wilson, Andrew Gordon and Saatchi, Yunus and Hu, Zhiting and Xing, Eric P},
  journal={Journal of Machine Learning Research},
  volume={18},
  number={82},
  pages={1--37},
  year={2017}
}

@book{williams2006gaussian,
  title={Gaussian processes for machine learning},
  author={Williams, Christopher KI and Rasmussen, Carl Edward},
  year={2006},
  publisher={MIT press Cambridge, MA}
}

@inproceedings{hensman2013gaussian,
  title={Gaussian Processes for Big Data},
  author={Hensman, James and Fusi, Nicol{\`o} and Lawrence, Neil D.},
  booktitle={Uncertainty in Artificial Intelligence},
  year={2013},
  volume={29},
  pages={282--290},
  publisher={AUAI Press},
  url={http://auai.org/uai2013/prints/papers/244.pdf}
}

@article{lyu2023efficient,
  title={Efficient bayesian optimization with deep kernel learning and transformer pre-trained on multiple heterogeneous datasets},
  author={Lyu, Wenlong and Hu, Shoubo and Chuai, Jie and Chen, Zhitang},
  journal={arXiv preprint arXiv:2308.04660},
  year={2023}
}

@article{huang2019clinicalbert,
  title={Clinicalbert: Modeling clinical notes and predicting hospital readmission},
  author={Huang, Kexin and Altosaar, Jaan and Ranganath, Rajesh},
  journal={arXiv preprint arXiv:1904.05342},
  year={2019}
}

@inproceedings{hensman2015scalable,
  title={Scalable variational Gaussian process classification},
  author={Hensman, James and Matthews, Alexander and Ghahramani, Zoubin},
  booktitle={Artificial intelligence and statistics},
  pages={351--360},
  year={2015},
  organization={PMLR}
}

@article{mikolov2013efficient,
  title={Efficient estimation of word representations in vector space},
  author={Mikolov, Tomas and Chen, Kai and Corrado, Greg and Dean, Jeffrey},
  journal={arXiv preprint arXiv:1301.3781},
  year={2013}
}

@article{hochreiter1997long,
  title={Long short-term memory},
  author={Hochreiter, Sepp and Schmidhuber, J{\"u}rgen},
  journal={Neural computation},
  volume={9},
  number={8},
  pages={1735--1780},
  year={1997},
  publisher={MIT press}
}

@article{cho2014learning,
  title={Learning phrase representations using RNN encoder-decoder for statistical machine translation},
  author={Cho, Kyunghyun and Van Merri{\"e}nboer, Bart and Gulcehre, Caglar and Bahdanau, Dzmitry and Bougares, Fethi and Schwenk, Holger and Bengio, Yoshua},
  journal={arXiv preprint arXiv:1406.1078},
  year={2014}
}

@article{li2023time,
  title={Time series as images: Vision transformer for irregularly sampled time series},
  author={Li, Zekun and Li, Shiyang and Yan, Xifeng},
  journal={Advances in Neural Information Processing Systems},
  volume={36},
  pages={49187--49204},
  year={2023}
}

@inproceedings{wen2023transformers,
  title={Transformers in time series: A survey},
  author={Wen, Qingsong and Zhou, Tian and Zhang, Chaoli and Chen, Weiqi and Ma, Ziqing and Yan, Junchi and Sun, Liang},
  booktitle={International Joint Conference on Artificial Intelligence(IJCAI)},
  year={2023}
}

@inproceedings{schulam2015clustering,
  title={Clustering longitudinal clinical marker trajectories from electronic health data: Applications to phenotyping and endotype discovery},
  author={Schulam, Peter and Wigley, Fredrick and Saria, Suchi},
  booktitle={Proceedings of the AAAI Conference on Artificial Intelligence},
  volume={29:1},
  pages={2956--2964},
  year={2015}
}

@article{che2018recurrent,
  title={Recurrent neural networks for multivariate time series with missing values},
  author={Che, Zhengping and Purushotham, Sanjay and Cho, Kyunghyun and Sontag, David and Liu, Yan},
  journal={Scientific reports},
  volume={8},
  number={1},
  pages={6085},
  year={2018},
  publisher={Nature Publishing Group}
}

@inproceedings{alsentzer2019publicly,
  title={Publicly available clinical BERT embeddings},
  author={Alsentzer, Emily and Murphy, John and Boag, William and Weng, Wei-Hung and Jindi, Di and Naumann, Tristan and McDermott, Matthew},
  booktitle={Proceedings of the 2nd Clinical Natural Language Processing Workshop},
  pages={72--78},
  year={2019}
}

@article{rasmy2021med,
  title={Med-BERT: pretrained contextualized embeddings on large-scale structured electronic health records for disease prediction},
  author={Rasmy, Laila and Xiang, Yang and Xie, Ziqian and Tao, Cui and Zhi, Degui},
  journal={NPJ digital medicine},
  volume={4},
  number={1},
  pages={86},
  year={2021},
  publisher={Nature Publishing Group}
}

@article{rajkomar2018scalable,
  title={Scalable and accurate deep learning with electronic health records},
  author={Rajkomar, Alvin and Oren, Eyal and Chen, Kai and Dai, Andrew M and Hajaj, Nissan and Hardt, Michaela and Liu, Peter J and Liu, Xiaobing and Marcus, Jake and Sun, Mimi and others},
  journal={NPJ digital medicine},
  volume={1},
  number={1},
  pages={18},
  year={2018},
  publisher={Nature Publishing Group}
}

@article{si2021deep,
  title={Deep representation learning of patient data from electronic health records (EHR): a systematic review},
  author={Si, Yuqi and Du, Jingcheng and Li, Zhao and Jiang, Xiaoqian and Miller, Timothy and Wang, Fei and Zheng, W Jim and Roberts, Kirk},
  journal={Journal of biomedical informatics},
  volume={115},
  pages={103671},
  year={2021},
  publisher={Elsevier}
}

@article{zhang2022improving,
  title={Improving medical predictions by irregular multimodal electronic health records modeling},
  author={Zhang, Xinlu and Xu, Shiyang and Zhang, Luoyao and Qiao, Jingran and Chen, Jing},
  journal={arXiv preprint arXiv:2210.12156},
  year={2022}
}

@article{karami2024tee4ehr,
  title={TEE4EHR: Transformer event encoder for better representation learning in electronic health records},
  author={Karami, Hojjat and Guo, Tyler and Svoboda, David and Argha, Ahmadreza and Mahadevan, Aditya and Dobson, Richard and Firouzi, Behraad},
  journal={Artificial Intelligence in Medicine},
  volume={153},
  pages={102909},
  year={2024},
  publisher={Elsevier}
}

@article{shukla2019modeling,
  title={Modeling irregularly sampled clinical time series},
  author={Shukla, Satya Narayan and Marlin, Benjamin M},
  journal={arXiv preprint arXiv:1906.04716},
  year={2019}
}

@article{feuerriegel2021analyzing,
  title={Analyzing patient trajectories with artificial intelligence},
  author={Feuerriegel, Stefan and Frauen, Dennis and Melnychuk, Valentyn and Schweisthal, Jonas and Hess, Konstantin and Curth, Alicia and Bauer, Stefan and Kilbertus, Niki and Papakonstantinou, Isaac and van der Schaar, Mihaela},
  journal={Journal of Medical Internet Research},
  volume={23},
  number={12},
  pages={e29812},
  year={2021},
  publisher={JMIR Publications Inc., Toronto, Canada}
}

@article{wang2022deep,
  title={Deep learning approaches for predicting glaucoma progression using electronic health records and natural language processing},
  author={Wang, Sally Yu-Yun and Tseng, Brian and Hernandez-Boussard, Tina},
  journal={Ophthalmology Science},
  volume={2},
  number={2},
  pages={100127},
  year={2022},
  publisher={Elsevier}
}

@article{yohannan2020assessing,
  title={Assessing glaucoma progression using machine learning trained on longitudinal visual field and clinical data},
  author={Yohannan, Joel and Boland, Michael V and Bolduc, Valentin and Sivak, Jacky M and Chen, Teodora Sajovic and Friedman, David S and Ramulu, Pradeep Y and De Moraes, Carlos Gustavo Vde and others},
  journal={Ophthalmology},
  volume={128},
  number={7},
  pages={1016--1026},
  year={2021},
  publisher={Elsevier}
}

@article{yang2022large,
  title={A large language model for electronic health records},
  author={Yang, Xi and Chen, Aokun and PourNejatian, Nima and Shin, Hoo Chang and Smith, Kaleb E and Parisien, Christopher and Compas, Colin and Martin, Cheryl and Costa, Anthony B and Flores, Mona G and others},
  journal={NPJ digital medicine},
  volume={5},
  number={1},
  pages={194},
  year={2022},
  publisher={Nature Publishing Group}
}

@article{wang2023prediction,
  title={Prediction models for glaucoma in a multicenter electronic health records consortium: The Sight Outcomes Research Collaborative},
  author={Wang, Sally Yu-Yun and Singh, Kuldev and Lin, Sophia C and Hernandez-Boussard, Tina and Ritch, Robert and Rosen, Leopold and Stein, Joshua D},
  journal={Ophthalmology Glaucoma},
  volume={7},
  number={2},
  pages={162--172},
  year={2024},
  publisher={Elsevier}
}

@article{placido2023deep,
  title={VaDeSC-EHR: a transformer-based variational autoencoder for clustering longitudinal survival data from electronic health records},
  author={Placido, Davide and Yuan, Boya and Hjaltelin, Jonas X and Zheng, Chunlei and Haue, Amalie D and Chmura, Piotr J and Brunak, Søren and others},
  journal={Nature Communications},
  volume={16},
  number={1},
  pages={1--14},
  year={2025},
  publisher={Nature Publishing Group}
}

@article{liu2019representation,
  title={Representation learning for clinical time series prediction tasks in electronic health records},
  author={Liu, Zhenyu and Chen, Chen and Li, Luyu and Zhou, Jiayu and Li, Xiaoqian and Song, Le and Qi, Yuan},
  journal={BMC Medical Informatics and Decision Making},
  volume={19},
  number={1},
  pages={1--14},
  year={2019},
  publisher={Springer}
}

@article{ling2023comprehensive,
  title={A comprehensive survey of deep learning for time series forecasting: architectural diversity and open challenges},
  author={Ling, Chen and Dong, Xujiang and Zhao, Yangli and Chen, Yucheng and Liu, Huan and Wei, Yanzhi},
  journal={Artificial Intelligence Review},
  volume={57},
  number={1},
  pages={1--47},
  year={2025},
  publisher={Springer}
}

@article{lim2021time,
  title={Time-series forecasting with deep learning: a survey},
  author={Lim, Bryan and Ar{\i}k, Sercan {\"O} and Loeff, Nicolas and Pfister, Tomas},
  journal={Philosophical Transactions of the Royal Society A},
  volume={379},
  number={2194},
  pages={20200209},
  year={2021},
  publisher={The Royal Society Publishing}
}

@inproceedings{ober2021promises,
  title={The promises and pitfalls of deep kernel learning},
  author={Ober, Sebastian W and Rasmussen, Carl E and van der Wilk, Mark},
  booktitle={Uncertainty in Artificial Intelligence},
  pages={1206--1216},
  year={2021},
  organization={PMLR}
}

@inproceedings{zhang2020self,
  title={Self-attentive Hawkes process},
  author={Zhang, Qiang and Lipani, Aldo and Kirnap, Omer and Yilmaz, Emine},
  booktitle={International conference on machine learning},
  pages={11183--11193},
  year={2020},
  organization={PMLR}
}

@inproceedings{zuo2020transformer,
  title={Transformer hawkes process},
  author={Zuo, Simiao and Jiang, Haoming and Li, Zichong and Zhao, Tuo and Zha, Hongyuan},
  booktitle={International conference on machine learning},
  pages={11692--11702},
  year={2020},
  organization={PMLR}
}

@inproceedings{
mei2022transformer,
title={Transformer Embeddings of Irregularly Spaced Events and Their Participants},
author={Hongyuan Mei and Chenghao Yang and Jason Eisner},
booktitle={International Conference on Learning Representations},
year={2022},
url={https://openreview.net/forum?id=Rty5g9imm7H}
}

@inproceedings{
dosovitskiy2021an,
title={An Image is Worth 16x16 Words: Transformers for Image Recognition at Scale},
author={Alexey Dosovitskiy and Lucas Beyer and Alexander Kolesnikov and Dirk Weissenborn and Xiaohua Zhai and Thomas Unterthiner and Mostafa Dehghani and Matthias Minderer and Georg Heigold and Sylvain Gelly and Jakob Uszkoreit and Neil Houlsby},
booktitle={International Conference on Learning Representations},
year={2021},
url={https://openreview.net/forum?id=YicbFdNTTy}
}

@article{shickel2017deep,
  title={Deep EHR: a survey of recent advances in deep learning techniques for electronic health record (EHR) analysis},
  author={Shickel, Benjamin and Tighe, Patrick James and Bihorac, Azra and Rashidi, Parisa},
  journal={IEEE journal of biomedical and health informatics},
  volume={22},
  number={5},
  pages={1589--1604},
  year={2017},
  publisher={IEEE}
}

@article{qiu2025deep,
  title={Deep representation learning for clustering longitudinal survival data from electronic health records},
  author={Qiu, Jiajun and Hu, Yao and Li, Li and Erzurumluoglu, Abdullah Mesut and Braenne, Ingrid and Whitehurst, Charles and Schmitz, Jochen and Arora, Jatin and Bartholdy, Boris Alexander and Gandhi, Shrey and others},
  journal={Nature Communications},
  volume={16},
  number={1},
  pages={2534},
  year={2025},
  publisher={Nature Publishing Group UK London}
}

@article{bommakanti2020application,
  title={Application of the sight outcomes research collaborative ophthalmology data repository for triaging patients with glaucoma and clinic appointments during pandemics such as COVID-19},
  author={Bommakanti, Nikhil K and Zhou, Yunshu and Ehrlich, Joshua R and Elam, Angela R and John, Denise and Kamat, Shivani S and Kelstrom, Jared and Newman-Casey, Paula Anne and Shah, Manjool M and Weizer, Jennifer S and others},
  journal={JAMA ophthalmology},
  volume={138},
  number={9},
  pages={974--980},
  year={2020},
  publisher={American Medical Association}
}

@book{fisher1995statistical,
  title={Statistical analysis of circular data},
  author={Fisher, Nicholas I},
  year={1995},
  publisher={cambridge university press}
}

@article{lundberg2017unified,
  title={A unified approach to interpreting model predictions},
  author={Lundberg, Scott M and Lee, Su-In},
  journal={Advances in neural information processing systems},
  volume={30},
  year={2017}
}

@article{de2019gru,
  title={GRU-ODE-Bayes: Continuous modeling of sporadically-observed time series},
  author={De Brouwer, Edward and Simm, Jaak and Arany, Adam and Moreau, Yves},
  journal={Advances in neural information processing systems},
  volume={32},
  year={2019}
}

@article{chen2018neural,
  title={Neural ordinary differential equations},
  author={Chen, Ricky TQ and Rubanova, Yulia and Bettencourt, Jesse and Duvenaud, David K},
  journal={Advances in neural information processing systems},
  volume={31},
  year={2018}
}

@article{kidger2020neural,
  title={Neural controlled differential equations for irregular time series},
  author={Kidger, Patrick and Morrill, James and Foster, James and Lyons, Terry},
  journal={Advances in neural information processing systems},
  volume={33},
  pages={6696--6707},
  year={2020}
}

@article{Leske2003,
   author = {Leske, M Cristina and Heijl, Anders and Hussein, Mohamed 
             and Bengtsson, Bo and Hyman, Leslie and Komaroff, Eugene 
             and {Early Manifest Glaucoma Trial Group}},
   title = {Factors for glaucoma progression and the effect of treatment: 
            the early manifest glaucoma trial},
   journal = {Archives of Ophthalmology},
   volume = {121},
   number = {1},
   pages = {48--56},
   year = {2003}
 }

@article{Leske2007,
   author = {Leske, M Cristina and Heijl, Anders and Hyman, Leslie 
             and Bengtsson, Boel and Dong, LiMing and Yang, Zhongming 
             and {EMGT Group}},
   title = {Predictors of long-term progression in the early manifest 
            glaucoma trial},
   journal = {Ophthalmology},
   volume = {114},
   number = {11},
   pages = {1965--1972},
   year = {2007}
 }

@article{DeMoraes2011,
   author = {De Moraes, Carlos Gustavo V and Juthani, Viral J 
             and Liebmann, Jeffrey M and Teng, Christopher C 
             and Tello, Celso and Susanna Jr, Remo and Ritch, Robert},
   title = {Risk factors for visual field progression in treated 
            glaucoma},
   journal = {Archives of Ophthalmology},
   volume = {129},
   number = {5},
   pages = {562--568},
   year = {2011}
 }
